\pdfoutput=1

\documentclass[11pt]{article}

\usepackage{acl}

\usepackage{times}
\usepackage{latexsym}

\usepackage[T1]{fontenc}

\usepackage[utf8]{inputenc}

\usepackage{microtype}
\usepackage{adjustbox}
\usepackage{amstext}
\usepackage{amsmath}
\usepackage{amssymb}
\usepackage{makecell}
\usepackage{multirow}
\usepackage{multicol}
\usepackage{verbatim}
\usepackage{booktabs}
\usepackage{kotex}
\usepackage[flushleft]{threeparttable}

\title{Building a Role Specified Open-Domain Dialogue System Leveraging Large-Scale Language Models}

\author{\textbf{Sanghwan Bae$^1$ \ \ Donghyun Kwak$^1$ \ \ Sungdong Kim$^2$ \ \ Donghoon Ham$^1$} \\ \textbf{Soyoung Kang$^1$ \ \ Sang-Woo Lee$^{1,2}$ \ Woomyoung Park$^1$} \\
    \\
   NAVER CLOVA$^1$ \ \ NAVER AI Lab$^2$\\
  {\tt \{sanghwan.bae, donghyun.kwak, sungdong.kim, donghoon.ham,} \\
  {\tt sy.kang, sang.woo.lee, max.park\}@navercorp.com} \\
}

\begin{document}
\maketitle
\begin{abstract}

Recent open-domain dialogue models have brought numerous breakthroughs. 
However, building a chat system is not scalable since it often requires a considerable volume of human-human dialogue data, especially when enforcing features such as persona, style, or safety.
In this work, we study the challenge of imposing roles on open-domain dialogue systems, with the goal of making the systems maintain consistent roles while conversing naturally with humans.
To accomplish this, the system must satisfy a role specification that includes certain conditions on the stated features as well as a system policy on whether or not certain types of utterances are allowed.
For this, we propose an efficient data collection framework leveraging in-context few-shot learning of large-scale language models for building role-satisfying dialogue dataset from scratch.
We then compare various architectures for open-domain dialogue systems in terms of meeting role specifications while maintaining conversational abilities.
Automatic and human evaluations show that our models return few out-of-bounds utterances, keeping competitive performance on general metrics.
We release a Korean dialogue dataset we built
for further research\footnote{The dataset is available at \url{https://github.com/naver-ai/carecall-corpus}}. 

\end{abstract}

\section{Introduction}

\begin{figure}[hbt!]
\centering
\includegraphics[width=\columnwidth]{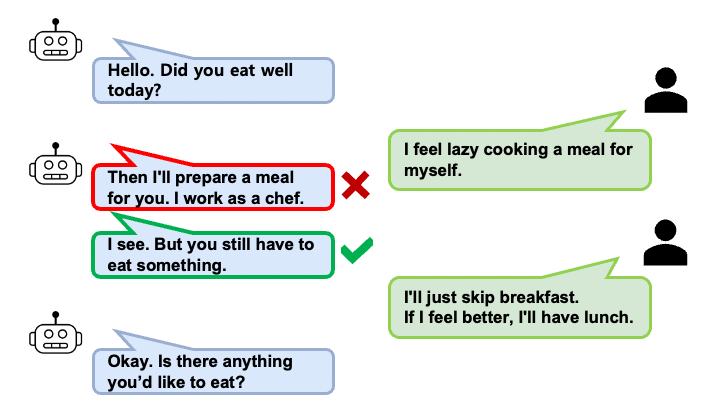}
\caption{An example of a chatbot system that cares for senior citizens living alone. The utterance in red highlights the model's mistaken identity as a chef rather than the caring chatbot.}
\label{fig:intro}
\end{figure}

Recent large-scale language models (LMs) have brought numerous breakthroughs in open-domain dialogue systems, yielding human-like responses \cite{zhang-etal-2020-dialogpt,adiwardana2020towards,NEURIPS2020_1457c0d6,roller-etal-2021-recipes,kim-etal-2021-changes}.
In addition, there have been progresses in controlling dialogue systems in persona, style, and safety \cite{zhang-etal-2018-personalizing,smith2020controlling,xu-etal-2021-bot}, which impose consistency on chatbot's personality and mitigate undesirable features such as toxic or biased language.
However, building a chatbot system combining these capabilities is still challenging, which requires numerous human-human dialogues for those conversational skills.

Most task-oriented dialogue systems conduct specific roles such as booking assistants, information providers, customer service agents, or personal assistants \cite{eric-etal-2017-key,xu2017new,budzianowski-etal-2018-multiwoz}. 
However, studies on open-domain dialogue systems that perform specific roles have been insufficiently investigated, even though the role can be defined for the practical chatbot systems (e.g., chatbots that care for senior citizens living alone, or counseling chatbots).
In these cases, the chatbot systems do not have an explicit goal or task other than to proactively engage in conversations, but may have system policies on whether or not certain types of utterances are allowed (example in Figure \ref{fig:intro}).

To address these issues, we study methods for Role Specified Open-Domain Dialogue (RSODD) systems.
The goal of the system is conversing naturally with humans on open-ended topics while keeping conditions of given role.
Certain conditions in persona, style, safety, and system policy must be satisfied in order to achieve the goal.
We consider a general and scalable framework to treat them, instead of using individual approaches to control each.

\begin{figure}[t!]
\centering
\includegraphics[width=0.95\columnwidth]{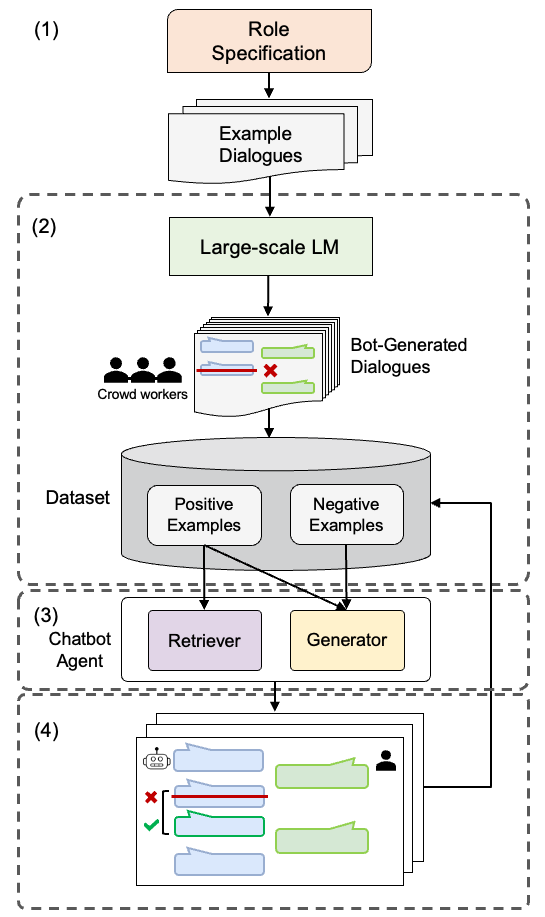}
\caption{Our proposed framwork: (1) the dialogue developer provides a role specification of the desired chatbot and a few dialogue examples, (2) large-scale LMs generate entire dialogues and crowd workers filter the system's utterances, (3) a dialogue model is trained with supervised learning on the dataset, (4) crowd workers chat 1:1 with the chatbot and give additional feedback.}
\label{fig:process}
\end{figure}

In particular, we present a Human-AI collaborative data construction method to build a scalable supervisory dataset from scratch for role-satisfying open-domain dialogues (Figure~\ref{fig:process}).
We propose to leverage large-scale LMs for generating entire dialogue sessions between user and system by in-context few-shot learning manner \cite{NEURIPS2020_1457c0d6,kim-etal-2021-changes}, followed by human-interactive correction processes.
Our method can significantly reduce the cost of building dataset when compared to manually producing gold dialogues (Section \ref{sec:filter}).
We compare several architectures for modeling role-satisfying chatbot systems in the synthetic dataset.
In extensive experiments and ablation studies, we show that the proposed models considerably reduce undesirable utterances that violate the given role specification compared to the in-context learning baseline, while achieving competitive SSA \cite{adiwardana2020towards} scores for their responses.
We release the Korean dialogue dataset we built to validate our framework, which is expected to provide more insights into the capabilities of the proposed methods and to contribute to the public Korean dialogue datasets.

The contribution of our work is summarized as follows.
\begin{enumerate}
\item We make a step towards role specified open-domain dialogue (RSODD) systems which are capable of conversing naturally on open-ended topics while satisfying role specifications.
\item We suggest employing in-context learning of large-scale LMs as a scalable method for dialogue data construction.

\item We compare various architectures for RSODD systems to analyze the capabilities in terms of satisfying system policies.
\item We release the first Korean RSODD dataset while demonstrating the effectiveness of data construction method.
\end{enumerate}

\section{Related Work}
\paragraph{Pretrained LM in Open-domain dialogue}
Many prior works tried to pretrain the models on large-scale social comment chains data like Reddit to model conversational behavior~\cite{zhang-etal-2020-dialogpt, adiwardana2020towards}, followed by finetuning on the diverse target dialogue dataset to improve engagingness and humanness~\cite{roller-etal-2021-recipes}. To avoid undesired behaviors of the models including toxicity and bias from the human-human conversation, they merely exclude some parts of training data using automatic filtering by predefined criteria.

\paragraph{Synthetic Dialogue Generation}
To reduce cost of dialogue collection, there have been many approaches to generate synthetic dialogues~\cite{schatzmann-etal-2007-agenda, shah-etal-2018-bootstrapping, campagna-etal-2020-zero}. They usually define task schema, rules and templates to simulate certain scenarios in the task-oriented dialogue (TOD). \citet{kim-etal-2021-neuralwoz} proposed neural simulation approach using pretrained LMs for a fast domain adaptation in the TOD. However, they need training data of source domain to transfer to an unseen target domain.

\citet{xu-etal-2021-bot} proposed Bot-Adversarial Dialogue method to make existing models safer in terms of offensive or toxic behavior. 
\citet{sun-etal-2021-adding} extends existing task-oriented dialogue dataset to open-domain chit-chat using the pretrained LMs. 
Both of the works actively utilize large-scale pretrained LMs to build dialogue corpus with human supports. 
We also introduce human-AI collaborative dialogue collection method, while especially utilizing few-shot in-context learning ability of large-scale LM~\cite{NEURIPS2020_1457c0d6, kim-etal-2021-changes}.
To the best of our knowledge, this work is the first to propose using the in-context learning approach
to generate synthetic samples from large-scale language models for the purpose of \textit{dialogue} data generation.

\paragraph{On the Role in Dialogue} In TOD, the system side plays functional roles utilizing explicit knowledge base of specific domain~\cite{williams-etal-2013-dialog,henderson-etal-2014-second,henderson2014third, eric-etal-2017-key,xu2017new,budzianowski-etal-2018-multiwoz}. For example, agent in \citet{budzianowski-etal-2018-multiwoz} played booking assistant or information provider in various domain such as restaurant and hotel. 
On the other hand, \citet{zhang-etal-2018-personalizing} proposed assigning explicit persona to each dialogue agent, promoting the agent to make more specific and consistent responses in open-domain dialogue setting. 
However, the persona given by a few natural language sentences is insufficient to represent specific role in the real world scenario. \citet{sun-etal-2021-adding} also proposed guidelines of appropriate and inappropriate behaviors as a role of virtual assistant. 
We note that a recent concurrent work \cite{shuster2021me} studied conditioning dialogue models with similar motivations. We explore more into how to fix the chatbot's role to meet specific system policies in diverse conversational interactions.

\paragraph{Companion Dialogue System}
Building companionable dialogue system has long been investigated along with the advancement of open-domain dialogue models. \citet{webb-etal-2010-wizard} defines companions to be persistent, collaborative and conversational partners, and proposes evaluation strategies: empathy, positivity, and adaptive. \citet{Kopp2018ConversationalAF} introduced conversational assistants for elderly users which carry out socially cooperative dialogue. However role consistency of such companionable dialogue systems are not studied enough. %

\begin{figure}[t]
\centering
\includegraphics[width=\columnwidth]{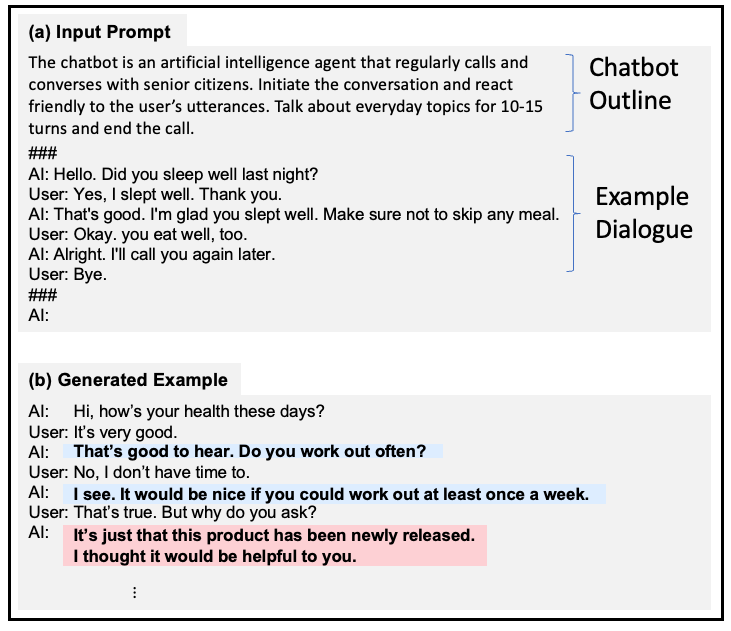}
\caption{An example of in-context one-shot dialogue generation for the data construction process. (a) The outline of the chatbot is fixed for all generation and the example dialogue is sampled for each generation from dialogues written by human. (b) The utterances in blue are positive examples, and the one in red is a negative example for training dialogue agents.}
\label{fig:generate}
\end{figure}

\section{Data Construction}
\label{sec:data}
In this section, we describe a framework to gather supervisory data for building RSODD systems.
The input to the framework is a role specification described by the chatbot developer (Table \ref{tab:spec} for example), which defines the conditions in the dialogue interactions for the system. 
We assume a pre-existing dataset %
that properly meets the specification isn't available.
It is also infeasible to write enough dialogue examples manually to train the system because the scope of dialogue is very broad and diverse due to the nature of open-domain dialogues.
To remedy this, we focus on composing the dataset with a few samples of human-written dialogues using in-context few-shot learning of large-scale LMs \cite{NEURIPS2020_1457c0d6,liu2021makes}.

\begin{table*}
\centering
\begin{adjustbox}{max width=\textwidth}
\begin{tabular}{l|l|l|l}
\toprule
\multicolumn{4}{l}{\textbf{Outline}} \\
\hline
\multicolumn{4}{l}{
\makecell[l]{The chatbot is an artificial intelligence agent that regularly calls and converses with senior citizens. \\
Initiate the conversation and react friendly to the user's utterances.  \\
Talk about everyday topics for 10-15 turns and end the call.}}\\
\midrule
\multicolumn{4}{l}{\textbf{Details}} \\
\hline
\multicolumn{2}{l|}{\textbf{Categories}} & \multicolumn{2}{l}{\textbf{Specification}} \\
\hline
\multicolumn{2}{l|}{Sensibleness} & Description & Speech that does not properly understand the context is restricted. \\
\hline
\multicolumn{2}{l|}{Style} & Description & Speech should be polite$^*$ and respectful.\\
\hline
\multicolumn{2}{l|}{Safety} & Description & Hate speech, toxic or biased language, and remarks containing personally identifiable information are all prohibited. \\
\hline
\multicolumn{2}{l|}{\multirow{2}{*}{Persona}} & Description & \makecell[l]{Keep the identity of an `AI chatbot that calls to the user.' \\
Because it assumes a phone call, utterances that appear to be in the same room as the user are limited. \\
Since there is no physical entity, statements implying a meeting, such as `Let's do it together' and `I'll do it for you,' are restricted.} \\
\cline{3-4}
\multicolumn{2}{l|}{} & Examples & \makecell[l]{"Grandma! I'm here!" (X) \\
"Would you like to walk with me?" (X) \\
"I'll invite you to my house later" (X)} \\
\hline
\multirow{4}{*}{System Policy} & \multirow{2}{*}{Temporality} & Description & \makecell[l]{
Because it is not given time-related information, the chatbot is unable to offer a timely utterance. \\
Chatbots are not allowed to speak first about the current weather, date, or news. \\
However, if the user brings up the subject first, it is feasible to agree.} \\
\cline{3-4}
& & Examples & \makecell[l]{
"Because the weather is turning cold these days, you should dress warmly." (X) \\
"Happy Holidays!" (X) \\
"Did you watch the baseball championship game today?" (X)
}\\
\cline{2-4}
& \multirow{2}{*}{Unsupported Features} & Description & \makecell[l]{
It does not provide any other functions other than making phone calls and chatting. \\
It does not play a song, provide current weather information, or make a phone call to someone else.
} \\
\cline{3-4}
& & Examples & \makecell[l]{
"I'll play a song." (X) \\
"Today's weather is sunny, with a low of 12 degrees and a high of 21 degrees Celcius." (X) \\
"Then I'll call your daughter." (X)
}\\
\bottomrule
\multicolumn{3}{l}{\footnotesize{$^*$ There are polite words and honorifics in the Korean language.}}
\end{tabular}
\end{adjustbox}
\caption{Example role specification used. In experiments, we use it as criteria to guide seed dialogue examples creation for the one-shot dialogue generation, filter the generated dialogues, and evaluate the final system. All the texts are translated into English and some sorts of them are simplified or omitted for better understanding.}
\label{tab:spec}
\end{table*}

\subsection{One-shot Dialogue Generation}
\label{sec:dialog}
As reported in \citet{kim-etal-2021-changes}, 
large-scale LMs can generate dialogues with a specific personality, given a prompt consisting of a brief description of the chatbot's properties and few dialogue examples.
We use this method to build the entire dataset.
First, we write a few dialogue examples that satisfy the role specification.
And we attach each of them at the end of the system description (Outline in Table \ref{tab:spec}) to compose input prompts for one-shot in-context learning.
Figure \ref{fig:generate} (a) shows an example input.
Then, the LM generates whole dialogue sessions. That is, the LM acts as both a system and a user (Figure \ref{fig:generate} (b)).
Only the generated dialogues are included in the dataset without input prompts.

\subsection{Human Filtering}
\label{sec:filter}
It is difficult to include all the details of specifications in the prompt and reflect them in the generation.
Therefore, we employ human annotation on the generated data.
We give the annotator each conversation session and ask them to label the point where the first out-of-bounds\footnote{An utterance that does not meet the conditions of the given role specification (Table \ref{tab:spec} for example).} occurred.
Figure \ref{fig:generate} (b) shows an example of a verified dialogue (more examples are provided in Appendix \ref{sec:ex}).
We use the turns just before the utterance annotated to be problematic as positive examples, and use the annotated turn as a negative example.
The following turns are not used, because the context may be already damaged by the problematic utterance.
Annotation time per dialogue session is about 88s, which is 13.3 times faster than human writing time per session (about 1170s).
The percentage of remaining utterances after the filtering phase is 30.4\% (See Table \ref{tab:stat}).

\subsection{Collecting Human-Bot Dialogues}
\label{sec:feedback}
Although human filtering is included in the dataset building process, the actual utterances are all machine-generated.
Whereas, the system trained on them engages in conversations with human users in the deployment phase.
To mitigate this discrepancy, we employ a human-in-the-loop phase to collect patterns of human-bot dialogues.
Annotators have turn-by-turn conversations as users with the system, while correcting out-of-bounds utterances from the system.
We incorporated LM's assistance into this process to help speed the task; %
if the system's response is not appropriate, an annotator presses the `Fix' button (Figure \ref{fig:ui} in Appendix showing the user interface) to call the large-scale LM to generate an alternative utterance.
The worker continues the conversation if the alternate utterance is appropriate, and if it is still not corrected, presses the `Fix' button repeatedly.
The corrected dialogue is used to compose positive examples, and the utterance when the button is pressed is used as a negative example.
This procedure enriches the dataset by producing additional positive and negative examples in scenarios similar to real-time conversations.

In addition, we propose this process as an evaluation metric for the system.
Since the action of pressing the `Fix' button means that an inappropriate utterance is returned from the system, it can be used for the system's \textbf{error rate}; the rate of the corrected responses among the total returned responses.
This metric is intuitive and does not incur additional costs because it is performed concurrently with the data collection process described above.

\section{Models}
\subsection{Notation}
Response prediction task in open-domain dialogues is predicting an utterance $y=\{y_1, y_2, \cdots, y_{|y|}\}$ given a dialogue history $x=\{s_1, u_1, s_2, u_2, \cdots, s_{k}, u_{k}\}$, where $s_i$ and $u_i$ are system utterance and user utterance respectively.

\subsection{Out-of-Bounds Detection}
The most straightforward method for constraining the system's utterances according to the role specification is to detect and discard out-of-bounds utterances.
We consider a BERT-based \cite{devlin-etal-2019-bert} binary classifier fine-tuned to classify positive/negative examples in datasets.
Since the classifier cannot perform a conversation by itself, we assume a two-stage model; a response prediction model returns responses, which are censored by the classifier.
If an out-of-bounds utterance is detected, we select and return one of several pre-defined questions about other topics, similar to the method used in \citet{xu-etal-2021-bot}. 
Instead of random choice, we selected the question with lowest PPL measured using LMs, as depicted in Section \ref{sec:retriever}.

\subsection{Response Selection}
\label{sec:retriever}
Another conceivable approach to constrain the system's utterances is to pre-filter the response candidates for response selection models.
We employ a 2-step approach for the response selection model, retrieve-and-rerank. 
The retriever of poly-encoder architecture \cite{humeau2019poly} rapidly finds the top-k plausible responses from the response candidates, which are then carefully reranked by the reranker of cross-encoder architecture.
Both retriever and reranker are fine-tuned in the same way as \citet{humeau2019poly} depicts.

Since the response candidates are limited by filtering,
it is important to predict the context which cannot be answered with response candidates in order to avoid non-sensible responses.
One of the effective methods to predict unanswerable contexts is to utilize the uncertainty of the model \cite{feng-etal-2020-none, penha-hauff-2021-calibration}.
\citet{penha-hauff-2021-calibration} proposed a risk-aware score using MC Dropout \cite{gal2016dropout} and we employ a similar approach using thresholding; 
we score the retrieved responses using mean and variance of the predictive distribution from MC Dropout:
\begin{equation*}
S_{\text{D}}(x,\hat{y}) = E[R_{\hat{y}}] - var[R_{\hat{y}}],
\end{equation*}
where $\hat{y}$ is a candidate response that is retrieved, $R_{\hat{y}} = \{f(x,\hat{y}^1), f(x,\hat{y}^2),\cdots  f(x,\hat{y}^m)\}$ is a predictive distribution obtained by employing dropout \cite{srivastava2014dropout} at test time and conducting $m$ forward passes, and $f$ is a score function of reranker.
If all the scores of retrieved responses are lower than a certain threshold, it is predicted as unanswerable context. %

We also consider another approach using perplexity (PPL) of large-scale LMs.
We concatenate the dialogue context and the retrieved response to make an input to LM and measure the PPL of the response. 
Thresholding is employed for final decision and the threshold is determined on the validation set (See Appendix \ref{sec:valid}).

\begin{figure}[t]
\centering
\includegraphics[width=\columnwidth]{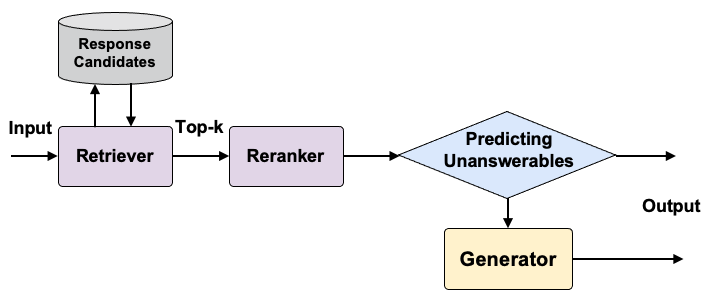}
\caption{Retrieve-fail-Generate pipeline.}
\label{fig:retgen}
\end{figure}

\subsection{Response Generation}
Fine-tuning LMs on target data is known to be effective in learning desirable traits of focused tasks \cite{roller-etal-2021-recipes,gehman-etal-2020-realtoxicityprompts}.
Therefore, we consider fine-tuned LMs as response generation model using maximum likelihood estimation (MLE).
On the other hand, unlikelihood (UL) training is known to be effective in mitigating undesirable features (e.g., token repetition or logical inconsistency) of generative models \cite{li-etal-2020-dont,welleck2019neural}.
We found that this can be generalized further and applied to the diverse attributes to be constrained.
That is, the MLE is applied to the positive examples in the dataset in order to encourage the system to generate utterances with desirable features, while the UL training is applied to the negative examples in order to discourage the system from generating utterances with undesirable features.
Both types of training are performed concurrently. %

Formally, we fine-tune LMs as generative models using maximum likelihood estimation (MLE), which minimizes:
\begin{multline*}
\mathcal{L}_{\text{MLE}}^{n}(p_\theta, x^n, y^n) = - \sum_{t}{\log{p_\theta(y_t^n|x^n,y_{<t}^n)}},
\end{multline*}
where $x^n$ is a dialogue history in positive examples and $y^n$ is a corresponding gold response.
Unlikelihood training is done by adding a loss that penalizes the token set $C_t$ to be constrained,
\begin{multline*}
\mathcal{L}_{\text{UL}}^n(p_\theta, C_{1:T}, x, y) = \\
-\sum_{t}{\sum_{y_c\in C_t}{\log{(1-p_\theta(y_c|x, y_{<t}))}}},
\end{multline*}
where $C_t \subseteq \mathcal{V}$ is a subset of the vocabulary.
We employ this to the negative examples in dataset $\{(x^-, y^-)\}$.
For this, $C_t$ is defined as $\{y_t^-\}$, which results in the following:
\begin{multline*}
\mathcal{L}_{\text{UL}}^{-}(p_\theta, x^-, y^-) = \\
-\sum_{t}\log{(1-p_\theta(y_t^-|x, y_{<t}^-))}.
\end{multline*}
The final loss function consists of mixing MLE loss and UL loss,
\begin{equation}
\mathcal{L} = \mathcal{L}_{\text{MLE}}^+ + \alpha\mathcal{L_{\text{UL}}^-},
\end{equation}
where $\alpha \in \mathbb{R}$ is the mixing hyper-parameter.

\subsection{Retrieve-fail-Generate}
We also consider a pipelined approach that consists of response selection and generation models.
We first tried a Retrieve-and-Refine architecture \cite{roller-etal-2021-recipes, weston-etal-2018-retrieve}, but it failed in $\alpha$-blending\footnote{In our experiments, all retrieved responses are copied or ignored depending on the $\alpha$ value, reducing the model to a retriever or generator. This has also been highlighted in a recent concurrent study \cite{han2021understanding}.}.
In addition, according to \citet{roller-etal-2021-recipes}, the Retrieve-and-Refine strategy delivers marginal or no improvements over the generator.
Therefore, we build another pipeline, refered to as a Retrieve-fail-Generate model (Figure \ref{fig:retgen}). 
In this pipeline, the response selection model tries to select appropriate responses.
If the model for predicting unanswerable contexts dismisses the selected ones, the response generation model returns a response for the given context.
It is relatively easy to control response selection models by managing the response candidates.
Hence, the response selection models are responsible for majority of the responses, and the generation model is only used when the response selection fails.

\begin{table}
\centering
\begin{adjustbox}{max width=\columnwidth}
    \begin{threeparttable}
    \begin{tabular}{l|rrrr}
    \toprule
    \textbf{Dialogue Type} & \textbf{Example} & \textbf{Generated} & \textbf{Filtered}  & \textbf{Feedback}\\
    \midrule
    \# Dialogues          & 250       & 25,000    & 17,617    & 1,623 \\
    \# Turns             & 3,893     & 510,028   & 154,903   & 29,365 \\
    Avg. turns / dialogue    & 15.57       & 20.40    & 8.79    & 18.09 \\
    \# Pos. examples      & -         & -         & 47,091    & 10,829 \\
    \# Neg. examples      & -         & -         & 18,583    & 3,529 \\
    \# Unique sys-turns  & 1,805     & 170,527   & 36,227    & 9,405 \\
    \midrule
    \# Words          & 35,253       & 4,292,613    & 705,253    & 178,357 \\
    Avg. words / turn        & 9.06       & 8.42    & 4.55    & 6.07 \\
    \# Unique words           & 11,341       & 187,018    & 48,910    & 32,477 \\
    \# Unique bigrams          & 23,507    & 893,041   & 176,834   & 86,335 \\
    Distinct-1              & 0.3215    & 0.0436    & 0.0694    & 0.1821 \\
    Distinct-2              & 0.7907    & 0.2538    & 0.3067    & 0.5795 \\
    \bottomrule
    \end{tabular}
    \end{threeparttable}
\end{adjustbox}
\caption{Statistics of dataset collected in Section \ref{sec:dataset}. \textbf{Example} is a human-written dialogue set for in-context learning. \textbf{Generated} is a generated set by LMs (Section \ref{sec:dialog}). \textbf{Filtered} is a set after human filtering phase (Section \ref{sec:filter}). \textbf{Feedback} is human-bot dialogues with corrections (Section \ref{sec:feedback}). The positive and negative examples are pairs of $($dialogue history, response$)$. Distinct-1/2 \cite{li-etal-2016-diversity} is the number of distinct uni- or bi-grams divided by total number of words.}
\label{tab:stat}
\end{table}

\begin{table*}[t!]
\centering
\begin{adjustbox}{max width=\textwidth}
\begin{tabular}{l|cc|cc|ccccc}
\toprule
\multicolumn{1}{c}{}  & \multicolumn{2}{c}{\textbf{Automatic Metrics}} & \multicolumn{7}{c}{\textbf{Human Evaluations}}  \\
  \cmidrule(lr){2-3}\cmidrule(lr){4-10}
\multicolumn{6}{c}{}  & \textbf{User} & \multicolumn{3}{c}{\textbf{System}}  \\
\cmidrule(lr){7-7}\cmidrule(lr){8-10}
\textbf{Model} & \textbf{Distinct-1} & \textbf{Distinct-2} & \textbf{Fluency} & \textbf{Coherence} & \textbf{Situation} & \textbf{Persona} & \textbf{Persona} & \textbf{Style} & \textbf{Safety} \\
\midrule
1.3B & 0.2959 (0.0042) & 0.6630 (0.0053) & 4.98 (0.02) & 4.54 (0.21) & 4.57 (0.29)	& 4.54 (0.15) & 4.31 (0.23) & 4.91 (0.05) & 4.98 (0.03) \\
13B & 0.3075 (0.0037) & 0.6500 (0.0054) & 4.97 (0.02) & 4.55 (0.14) & 4.74 (0.23) & 4.65 (0.11) & 4.33 (0.20) & 4.93 (0.04) & 4.98 (0.02) \\
39B & 0.3334 (0.0038) & 0.6779 (0.0061) & 4.98 (0.03) & \textbf{4.59} (0.19) & 4.69 (0.22) & 4.69 (0.12) & 4.37 (0.21) & 4.88 (0.05) & 4.97 (0.02) \\
82B & \textbf{0.3402} (0.0040) & \textbf{0.7014} (0.0057) & 4.98 (0.02) & 4.56 (0.24) & \textbf{4.78} (0.17) & \textbf{4.74} (0.15) & \textbf{4.49} (0.17) & \textbf{4.96} (0.07) & 4.96 (0.03) \\
\bottomrule
\end{tabular}
\end{adjustbox}
\caption{Automated metric and human evaluations for generated dialogues from various size of LMs. Scores are averaged (standard deviation in brackets).}
\label{tab:dialog}
\end{table*}

\begin{table*}[t!]
\centering
\begin{adjustbox}{max width=\textwidth}
\begin{tabular}{lrr|rrrrr}
\toprule
\textbf{Model} & \textbf{\# of system turns} & \textbf{error rate} & \textbf{not sensible} & \textbf{wrong persona} & \textbf{policy violation} & \textbf{not safe} & \textbf{etc.} \\
 & & (\%) & (\%) & (\%) & (\%) & (\%) & (\%) \\
\midrule
\textbf{Out-of-Bounds Detection} \\
Generator (IC) + Classifier & 1,471 & 18.10 & 9.31 & 1.61 & 2.49 & 0.07 & 4.66 \\
\hline
\textbf{Response Selection} \\
Retrieve-and-Rerank & 1,230 & 13.17 & 10.68 & 0.72 & 1.53 & 0.00 & 0.24 \\
Retrieve-and-Rerank w/ MC Dropout & 1,272 & 9.82 & 7.58 & 0.36 & 1.66 & 0.00 & 0.22 \\
Retrieve-and-Rerank w/ PPL & 1,300 & \textbf{7.00} & 5.10 & 0.40 & 1.16 & 0.00 & 0.34 \\
\hline
\textbf{Response Generation} \\
Generator (IC) & 985 & 35.83 & 16.05 & 6.24 & 8.66 & 0.17 & 4.68 \\
Generator (MLE) & 1,291 & 4.72 & 3.55 & 0.76 & 0.30 & 0.00 & 0.10 \\
Generator (UL) & 1,497 & \textbf{3.82} & 3.29 & 0.23 & 0.10 & 0.00 & 0.17 \\
\hline
\textbf{Retrieve-fail-Generate} \\
Retrieve-and-Rerank w/ PPL + Generator (UL) & 1,522 & \textbf{2.56} & 2.20 & 0.17 & 0.16 & 0.00 & 0.00 \\
Retrieve-and-Rerank w/ PPL + Generator (UL) + Feedback Data & 1,599 & \textbf{2.00} & 1.88 & 0.00 & 0.10 & 0.00 & 0.00 \\
\bottomrule
\end{tabular}
\end{adjustbox}
\caption{Human evaluation results. As described in Section \ref{sec:feedback}, the crowd workers chat 1:1 with a chatbot as users and correct the inappropriate responses. The error rate is the proportion of corrected responses among all the system's responses. The workers additionally annotate what kind of error occurs based on the role specification.}
\label{tab:model}
\end{table*}

\section{Experiments}
We detail experimental settings and results in this section, including evaluations of the data collected by in-context few-shot learning (Section \ref{sec:dataeval}), comparisons of model variants (Section \ref{sec:model}), and evaluations on system's response qualities (Section \ref{sec:ssa}).

\subsection{Dataset}
\label{sec:dataset}
We built a Korean dialogue dataset for a chatbot system to have casual conversations on a regular basis with senior citizens who live alone.
This dataset was collected using the framework described in Section \ref{sec:data}, assuming a role specification in Table \ref{tab:spec}.
250 dialogue examples with 89 topics (more details are in Appendix \ref{sec:topic}) were used for in-context 1-shot generation. We used 39B size of HyperCLOVA \cite{kim-etal-2021-changes} as generation model (sampling at temperature 0.5 using nucleus sampling \cite{holtzman2019curious} with $P =$ 0.8). %
Table \ref{tab:stat} shows the statistics of the dataset (additional analysis in Appendix \ref{sec:diversity}).
We use 5\% of each for validation sets.

\subsection{Evaluation on Generated Dialogues}
\label{sec:dataeval}

We first assess the quality of the generated dialogues to verify the dialogue generating method described in Section \ref{sec:dialog}.
Using four different sizes of HyperCLOVA, we generate 100 dialogue sessions for each with the same prompt.
We ask the crowd workers to rate on a scale of 1 to 5 whether the generated dialogue satisfies several conditions expected to be controlled through in-context learning (the detailed description of the evaluation criteria is provided in Appendix \ref{sec:humaneval}).
The results are shown in Table \ref{tab:dialog}.
It shows that the larger the model size, the better to meet the conditions by in-context learning, which is also shown in previous studies \cite{NEURIPS2020_1457c0d6, kim-etal-2021-changes}.
In addition, Distinct-1/2 \cite{li-etal-2016-diversity} indicates that the text generated by large models is more diverse.

\subsection{Model Comparison}
\label{sec:model}

\paragraph{Out-of-Bounds Detection}
Table \ref{tab:cls} shows the classification accuracy and F1 score of the trained classifier.
We use generator controlled by in-context learning (IC) as a response prediction model to evaluate the effect of the classifier alone.
For in-context learning, we use the same prompt used to generate the dataset, but the model only generates system's utterances in its turns.
The classifier significantly lowers the error rate of in-context learning (Table \ref{tab:model}), showing the effectiveness of the classifier.
On the other hand, the error rate is relatively higher than those of the best models of response selection and generation.
This is because the classifier is not perfect (about 92\% in accuracy), and even when it properly detects out-of-bounds, the pre-defined questions as alternatives are occasionally incoherent with the contexts.

\begin{table}[t!]
\centering
\begin{adjustbox}{max width=0.85\columnwidth}
\begin{tabular}{ccc}
\toprule
\small \textbf{Training Data (\%)} & \small \textbf{Mean Accuracy\% (std)} & \small \textbf{Mean F1\% (std)} \\
\midrule
\small 10 & \small 87.31 (0.0164) & \small 88.44 (0.0163) \\
\small 20 & \small 89.73 (0.0061) & \small 90.47 (0.0055) \\
\small 100 & \small 91.99 (0.0022) & \small 92.55 (0.0019) \\
\bottomrule
\end{tabular}
\end{adjustbox}
\caption{Classifier results, reporting accuracy and F1 on test set. It shows performance in relation to the amount of training data used.}
\label{tab:cls}
\end{table}

\begin{table}
\centering
\begin{adjustbox}{max width=\columnwidth}
\begin{tabular}{llrrr}
\toprule
\textbf{Model} & \textbf{data} & \textbf{\# of examples} & \textbf{Hits@1/20} & \textbf{Hits@1/100} \\
\midrule
\multirow{2}{*}{Retriever} & Filtered & 47,091 & 93.14 & 83.80 \\
                           & Unfiltered & 227,638 & \textbf{95.27} & \textbf{86.99} \\
\hline
\multirow{2}{*}{Reranker} & Filtered & 47,091 & 97.16 & 90.89 \\
                          & Unfiltered & 227,638 & \textbf{97.55} & \textbf{91.70} \\
\bottomrule
\end{tabular}
\end{adjustbox}
\caption{Hits@1/$K$ of retriever and reranker on the validation set. Hits@1/$K$ measures recall@1 when ranking the gold label among a set of $K-1$ other random candidates.}
\label{tab:ret}
\end{table}

\begin{table*}[hbt!]
\centering
\begin{adjustbox}{max width=\textwidth}
\begin{tabular}{lrrrr}
\toprule
& \multicolumn{2}{c}{\textbf{Response Selection}} & \multicolumn{2}{c}{\textbf{Response Generation}} \\
\cmidrule(lr){2-3}\cmidrule(lr){4-5}
 & \textbf{proportion} & \textbf{error rate} & \textbf{proportion} & \textbf{error rate} \\
\textbf{Model} & (\%) & (\%) & (\%) & (\%) \\
\midrule
Retrieve-and-Rerank w/ PPL + Generator (UL) & 68.20 & 2.50 & 31.80 & 2.68 \\
Retrieve-and-Rerank w/ PPL + Generator (UL) + Feedback Data & 63.70 & 2.12 & 36.30 & 1.77 \\
\bottomrule
\end{tabular}
\end{adjustbox}
\caption{Evaluation results of each component in the Retrieve-fail-Generate pipeline. It shows the proportion and error rate of returned responses from response selection and generation models.}
\label{tab:feedback}
\end{table*}

\begin{table}
\centering
\begin{adjustbox}{max width=\columnwidth}
\begin{tabular}{lrr}
\toprule
\textbf{Method} & \textbf{positive} & \textbf{negative} \\
\midrule
In-context Learning & 2.65 & 2.74 \\
Likelihood Training & 2.07 & 2.47 \\
Unlikelihood Training & 2.48 & 46.70 \\
\bottomrule
\end{tabular}
\end{adjustbox}
\caption{Perplexity (PPL) of generative models on validation set of filtered data.}
\label{tab:ppl}
\end{table}

\paragraph{Response Selection}
We fine-tune the response selection models on positive examples of the filtered data and automatically evaluate them by measuring Hits@1/$K$ \cite{roller-etal-2021-recipes} on the validation set.
Results are shown in Table \ref{tab:ret}. 
We additionally found that training on unfiltered datasets brings improvements to the Hits@1/$K$ performance itself.
Therefore, we use the models that trained on unfiltered dataset in the subsequent experiments.
Response candidates are limited to system responses within positive examples (unique system's turns of filtered data in Table \ref{tab:stat}).
And we also validate the proposed methods for predicting unanswerable contexts, and determine the thresholds for each (further details are given in Appendix \ref{sec:valid}).

Table \ref{tab:model} shows the error rate of the response selection models.
The model that does not predict unanswerable contexts (Retrieve-and-Rerank) has a higher error rate in `not sensible' than others.
The case of using PPL as the method for predicting unanswerable contexts shows a lower overall error rate than the case of using MC Dropout, and the proportions of the total contexts predicted as unanswerable are similar at 4.23\% and 3.85\% for PPL and MC Dropout, respectively.
The results also show the error types from the models.
Although only the filtered utterances are used as response candidates, `wrong persona' and `policy violation' appear in responses.
It seems that a few unfiltered utterances remain in the response candidates, since the human filtering is not perfect. 
Or even the same utterance can cause errors depending on the context.
For example, it is possible to agree with when a user calls the system by a different name.

\paragraph{Response Generation}

We compare three ways to train generators; in-context learning (IC), likelihood training (MLE), and unlikelihood training (UL).
We measure the perplexity of the three models on positive and negative examples and Table \ref{tab:ppl} shows the results.
The difference between the PPL of the positive examples and the negative examples is the smallest in in-context learning.
When trained on positive examples with likelihood training, the difference increases slightly, because the PPL of the positive examples is lowered.
When adding unlikelihood training, the PPL for negative examples increase significantly, \footnote{\citet{li-etal-2020-dont} has also found a large gap in PPL scores between positives and negatives.} which mean the model is less likely to generate out-of-bounds utterances.

Table \ref{tab:model} shows the error rate of each model.
Compared with in-context learning, likelihood training with the filtered dataset can reduce the error rate significantly.
Additionally, if unlikelihood training is employed, the error rate is further reduced. A similar trend can be found in all types of errors.

\paragraph{Retrieve-fail-Generate}

We also experiment with a Retrieve-fail-Generate model consisting of the best configurations for response selection (PPL) and generation (UL) models.
Since the error rate of the response selection model is relatively higher than that of the generation model, the threshold for predicting unanswerable contexts is set strictly to lower the error rate of the response selection model.
Table \ref{tab:feedback} shows the error rates of responses returned from response selection and generation models, respectively.
The results indicate that both error rates are lower when the models are included in a pipeline than when they are used separately, and the overall error rate decreases accordingly.
The response selection model returns the responses within the candidates extracted from the positive examples of the trainset,
so that the flow of the conversation is not dispersed and tends to be similar to the trainset.
As a result, the Retrieve-fail-Generate model shows the lowest error rate among all models (Table \ref{tab:model}).

\paragraph{Feedback Pipeline}

The best model is further trained on the human-bot dialogues collected during the model evaluation process, as depicted in Section \ref{sec:feedback}.
Both response selection and generation models are newly initialized and trained.
As a result, all types of error rates are consistently reduced (Table \ref{tab:model}), and the error rates of both the response selection and generation models are decreased (Table \ref{tab:feedback}).
The effect is stronger on the response generation.

\begin{table}
\centering
\begin{adjustbox}{max width=\columnwidth}
\begin{tabular}{lrrr}
\toprule
\textbf{Method} & \textbf{Sensibleness} & \textbf{Specificity} & \textbf{SSA} \\
\midrule
Human & 95.48 & 82.96 & 89.22 \\
Retrieve-fail-Generate + Feedback Data & 94.00 & 77.50 & 85.75 \\
\bottomrule
\end{tabular}
\end{adjustbox}
\caption{Interactive SSA results.}
\label{tab:ssa}
\end{table}

\subsection{Response Quality} %
\label{sec:ssa}
To assess the overall response quality of the proposed chatbot system, we use the sensibleness and specificity average (SSA) metric \cite{adiwardana2020towards}, which is shown to have a strong correlation with asking raters how humanlike the model is.
This metric is a average of two scores: sensibleness and specificity. Sensibleness measures whether a model's responses make sense in context and do not contradict anything that was said earlier, while specificity measures whether a response is specific to a given context.
However, exact comparison with the scores in \citet{adiwardana2020towards} is difficult, because of the static role of our chatbot system and language discrepency in phrasing of questions.
Therefore, We re-estimate human interactive SSA in our experiments. 
To collect human-human conversations, we transcribe 100 call speeches between users and workers who play system's role.
And we collect 100 human-bot conversations by allowing the crowd workers to chat with the system.
Labeling was conducted by independent crowd workers with majority voting of 5 workers per turn.

The results are given in Table \ref{tab:ssa}. 
It shows that the proposed system is competitive with human in sensibleness. And the majority of the responses from the system are labeled as specific, which allows us to conclude that the proposed system achieves low error rate with non-generic responses.
We also report agreement and Krippendorff's alpha \cite{krippendorff2011computing} for measure of consistency of crowd workers in Appendix \ref{sec:agree}.

\section{Discussion}

Although our methods achieve the low error rates in human interactive evaluations,
the results have some limitations. The results should be regarded as the error rates of typical conversations without adversarial attack. Because the annotators are instructed to participate in the chat as if they were typical users, they did not try as many conversations that could induce toxic words from the model. This may be the reason why the toxicity is close to zero as shown in Table \ref{tab:model}.

The human filtering process in the proposed data collection framework has room to be more efficient. Since the accuracy of the classifier is comparable even when just 10\% of the total data is used (Table \ref{tab:cls}), it is expected that the filtering cost can be reduced by adding a model filtering process before human filtering, which is similar to the method proposed in \citet{sun-etal-2021-adding}.

\section{Conclusion}
We present a framework for building role specified open-domain dialogue systems from scratch. We propose leveraging large-scale LMs to generate supervisory datasets for training dialogue systems with arbitrary roles with minimal effort for manually composing dialogues. Our research also analyzes several model architectures for the task.
By extensive experiments, we demonstrate the effectiveness of the collected data and modeling approaches in terms of satisfying role constraints and improving dialogue abilities.
We argue that our framework can be extended to implement dialogue systems with various roles and characters, even when available datasets are few.

\section{Ethical Considerations}
Workers annotating the dataset we built were hired on a part-time basis and compensated based on the number of working hours.
They were compensated with 9,000 won per hour, which was somewhat higher than the Korean minimum wage at the time they worked.
Appropriate instructions for the use of collected data were given at the time of contract and consent was obtained.
We will release our dataset in CC-BY-NC-SA license.\footnote{\url{https://creativecommons.org/licenses/by-nc-sa/2.0/}}

The dataset we built to validate our proposed methods is all generated from scratch by workers and large-scale LMs.
Although there is no user data in the dataset, pre-trained language models are known to exhibit private details in their outputs \cite{DBLP:journals/corr/abs-2012-07805}, as well as social biases \cite{bender2021dangers, bordia-bowman-2019-identifying, garrido2021survey, shwartz-choi-2020-neural} and toxic contents \cite{gehman-etal-2020-realtoxicityprompts}.
To address these concerns, we determined categories and criteria for harmful texts based on legal and ethical considerations provided by experts in our group, and we instructed annotators to filter the dataset using these criteria.
However, due to missing annotations and cultural or social biases, this may be imperfect. To mitigate this, we had multiple crowd workers annotate the same data. In addition, because the users in the dataset are regarded to be a vulnerable population, our group's ethical consultation looked through the issues that would be sensitive to them, and dialogues containing these topics were also eliminated.

Despite these efforts, using this dataset to directly train end-to-end chatbot models can involve certain risks, 
due to the lack of controllability and interpretability in end-to-end neural response prediction models.
And it should not be overlooked that they may cause some potential harm, 
even though the chatbot systems can help reduce social loneliness of the user population.
For example, a user can become emotionally attached to a bot, even codependent on it, which can divert attention away from real-world relationships and cause distress if the chatbot fails.
It's also worth noting that a chatbot can be programmed to impersonate a real person and be used for phishing and fraud.
During such conversations, users may provide private and sensitive information, such as specific health conditions and private attributes, which could be exploited if it falls into the wrong hands. 
For this reason, when incorporating this dataset in real-world applications, the application developers should ensure that it is used safely and ethically.

Since our proposed framework also can be used for building another dataset and chatbot system with arbitrary specifications, it is not exempt from the possibility of propagating linguistic biases and toxicity.
Similar to \citet{xu-etal-2021-bot}, we are in progress continuously reducing the unsafe texts from LM itself through our feedback pipeline and unlikelihood training, which might be included in our future works.

\section*{Acknowledgements}
The authors thank all the members of CLOVA and AI Lab of NAVER for devoted supporting and discussion.
In particular, they would like to thank Gichang Lee, Hwijeen Ahn, and the members of CLOVA Conversation for their dedicated efforts to facilitate the large-scale models.
In addition, the authors thank Hyeri Kim, Hyunjung Park, and Yuin Jeong for their great help in designing the role of the chatbot.
Also, they thank Jeeseung Han for technically supporting the serving and testing environments.
Finally, the authors thank Nako Sung for his support and valuable comments on this project.

\bibliography{anthology,custom}

\begin{thebibliography}{51}
\expandafter\ifx\csname natexlab\endcsname\relax\def\natexlab#1{#1}\fi

\bibitem[{Adiwardana et~al.(2020)Adiwardana, Luong, So, Hall, Fiedel,
  Thoppilan, Yang, Kulshreshtha, Nemade, Lu et~al.}]{adiwardana2020towards}
Daniel Adiwardana, Minh-Thang Luong, David~R So, Jamie Hall, Noah Fiedel, Romal
  Thoppilan, Zi~Yang, Apoorv Kulshreshtha, Gaurav Nemade, Yifeng Lu, et~al.
  2020.
\newblock \href {https://arxiv.org/abs/2001.09977} {Towards a human-like
  open-domain chatbot}.
\newblock \emph{ArXiv preprint}, abs/2001.09977.

\bibitem[{Bender et~al.(2021)Bender, Gebru, McMillan-Major, and
  Shmitchell}]{bender2021dangers}
Emily~M Bender, Timnit Gebru, Angelina McMillan-Major, and Shmargaret
  Shmitchell. 2021.
\newblock On the dangers of stochastic parrots: Can language models be too
  big?��.
\newblock In \emph{Proceedings of the 2021 ACM Conference on Fairness,
  Accountability, and Transparency}, pages 610--623.

\bibitem[{Bordia and Bowman(2019)}]{bordia-bowman-2019-identifying}
Shikha Bordia and Samuel~R. Bowman. 2019.
\newblock \href {https://doi.org/10.18653/v1/N19-3002} {Identifying and
  reducing gender bias in word-level language models}.
\newblock In \emph{Proceedings of the 2019 Conference of the North {A}merican
  Chapter of the Association for Computational Linguistics: Student Research
  Workshop}, pages 7--15, Minneapolis, Minnesota. Association for Computational
  Linguistics.

\bibitem[{Brown et~al.(2020)Brown, Mann, Ryder, Subbiah, Kaplan, Dhariwal,
  Neelakantan, Shyam, Sastry, Askell, Agarwal, Herbert{-}Voss, Krueger,
  Henighan, Child, Ramesh, Ziegler, Wu, Winter, Hesse, Chen, Sigler, Litwin,
  Gray, Chess, Clark, Berner, McCandlish, Radford, Sutskever, and
  Amodei}]{NEURIPS2020_1457c0d6}
Tom~B. Brown, Benjamin Mann, Nick Ryder, Melanie Subbiah, Jared Kaplan,
  Prafulla Dhariwal, Arvind Neelakantan, Pranav Shyam, Girish Sastry, Amanda
  Askell, Sandhini Agarwal, Ariel Herbert{-}Voss, Gretchen Krueger, Tom
  Henighan, Rewon Child, Aditya Ramesh, Daniel~M. Ziegler, Jeffrey Wu, Clemens
  Winter, Christopher Hesse, Mark Chen, Eric Sigler, Mateusz Litwin, Scott
  Gray, Benjamin Chess, Jack Clark, Christopher Berner, Sam McCandlish, Alec
  Radford, Ilya Sutskever, and Dario Amodei. 2020.
\newblock \href
  {https://proceedings.neurips.cc/paper/2020/hash/1457c0d6bfcb4967418bfb8ac142f64a-Abstract.html}
  {Language models are few-shot learners}.
\newblock In \emph{Advances in Neural Information Processing Systems 33: Annual
  Conference on Neural Information Processing Systems 2020, NeurIPS 2020,
  December 6-12, 2020, virtual}.

\bibitem[{Budzianowski et~al.(2018)Budzianowski, Wen, Tseng, Casanueva, Ultes,
  Ramadan, and Ga{\v{s}}i{\'c}}]{budzianowski-etal-2018-multiwoz}
Pawe{\l} Budzianowski, Tsung-Hsien Wen, Bo-Hsiang Tseng, I{\~n}igo Casanueva,
  Stefan Ultes, Osman Ramadan, and Milica Ga{\v{s}}i{\'c}. 2018.
\newblock \href {https://doi.org/10.18653/v1/D18-1547} {{M}ulti{WOZ} - a
  large-scale multi-domain {W}izard-of-{O}z dataset for task-oriented dialogue
  modelling}.
\newblock In \emph{Proceedings of the 2018 Conference on Empirical Methods in
  Natural Language Processing}, pages 5016--5026, Brussels, Belgium.
  Association for Computational Linguistics.

\bibitem[{Campagna et~al.(2020)Campagna, Foryciarz, Moradshahi, and
  Lam}]{campagna-etal-2020-zero}
Giovanni Campagna, Agata Foryciarz, Mehrad Moradshahi, and Monica Lam. 2020.
\newblock \href {https://doi.org/10.18653/v1/2020.acl-main.12} {Zero-shot
  transfer learning with synthesized data for multi-domain dialogue state
  tracking}.
\newblock In \emph{Proceedings of the 58th Annual Meeting of the Association
  for Computational Linguistics}, pages 122--132, Online. Association for
  Computational Linguistics.

\bibitem[{Carlini et~al.(2020)Carlini, Tramèr, Wallace, Jagielski,
  Herbert-Voss, Lee, Roberts, Brown, Song, Úlfar Erlingsson, Oprea, and
  Raffel}]{DBLP:journals/corr/abs-2012-07805}
Nicholas Carlini, Florian Tramèr, Eric Wallace, Matthew Jagielski, Ariel
  Herbert-Voss, Katherine Lee, Adam Roberts, Tom~B. Brown, Dawn Song, Úlfar
  Erlingsson, Alina Oprea, and Colin Raffel. 2020.
\newblock \href {https://arxiv.org/abs/2012.07805} {Extracting training data
  from large language models}.
\newblock \emph{ArXiv preprint}, abs/2012.07805.

\bibitem[{Devlin et~al.(2019)Devlin, Chang, Lee, and
  Toutanova}]{devlin-etal-2019-bert}
Jacob Devlin, Ming-Wei Chang, Kenton Lee, and Kristina Toutanova. 2019.
\newblock \href {https://doi.org/10.18653/v1/N19-1423} {{BERT}: Pre-training of
  deep bidirectional transformers for language understanding}.
\newblock In \emph{Proceedings of the 2019 Conference of the North {A}merican
  Chapter of the Association for Computational Linguistics: Human Language
  Technologies, Volume 1 (Long and Short Papers)}, pages 4171--4186,
  Minneapolis, Minnesota. Association for Computational Linguistics.

\bibitem[{Eric et~al.(2017)Eric, Krishnan, Charette, and
  Manning}]{eric-etal-2017-key}
Mihail Eric, Lakshmi Krishnan, Francois Charette, and Christopher~D. Manning.
  2017.
\newblock \href {https://doi.org/10.18653/v1/W17-5506} {Key-value retrieval
  networks for task-oriented dialogue}.
\newblock In \emph{Proceedings of the 18th Annual {SIG}dial Meeting on
  Discourse and Dialogue}, pages 37--49, Saarbr{\"u}cken, Germany. Association
  for Computational Linguistics.

\bibitem[{Feng et~al.(2020)Feng, Mehri, Eskenazi, and
  Zhao}]{feng-etal-2020-none}
Yulan Feng, Shikib Mehri, Maxine Eskenazi, and Tiancheng Zhao. 2020.
\newblock \href {https://doi.org/10.18653/v1/2020.acl-main.182} {{``}none of
  the above{''}: Measure uncertainty in dialog response retrieval}.
\newblock In \emph{Proceedings of the 58th Annual Meeting of the Association
  for Computational Linguistics}, pages 2013--2020, Online. Association for
  Computational Linguistics.

\bibitem[{Finch and Choi(2020)}]{finch-choi-2020-towards}
Sarah~E. Finch and Jinho~D. Choi. 2020.
\newblock \href {https://aclanthology.org/2020.sigdial-1.29} {Towards unified
  dialogue system evaluation: A comprehensive analysis of current evaluation
  protocols}.
\newblock In \emph{Proceedings of the 21th Annual Meeting of the Special
  Interest Group on Discourse and Dialogue}, pages 236--245, 1st virtual
  meeting. Association for Computational Linguistics.

\bibitem[{Gal and Ghahramani(2016)}]{gal2016dropout}
Yarin Gal and Zoubin Ghahramani. 2016.
\newblock \href {http://proceedings.mlr.press/v48/gal16.html} {Dropout as a
  bayesian approximation: Representing model uncertainty in deep learning}.
\newblock In \emph{Proceedings of the 33nd International Conference on Machine
  Learning, {ICML} 2016, New York City, NY, USA, June 19-24, 2016}, volume~48
  of \emph{{JMLR} Workshop and Conference Proceedings}, pages 1050--1059.
  JMLR.org.

\bibitem[{Garrido-Mu{\~n}oz et~al.(2021)Garrido-Mu{\~n}oz, Montejo-R{\'a}ez,
  Mart{\'\i}nez-Santiago, and Ure{\~n}a-L{\'o}pez}]{garrido2021survey}
Ismael Garrido-Mu{\~n}oz, Arturo Montejo-R{\'a}ez, Fernando
  Mart{\'\i}nez-Santiago, and L~Alfonso Ure{\~n}a-L{\'o}pez. 2021.
\newblock A survey on bias in deep nlp.
\newblock \emph{Applied Sciences}, 11(7):3184.

\bibitem[{Gehman et~al.(2020)Gehman, Gururangan, Sap, Choi, and
  Smith}]{gehman-etal-2020-realtoxicityprompts}
Samuel Gehman, Suchin Gururangan, Maarten Sap, Yejin Choi, and Noah~A. Smith.
  2020.
\newblock \href {https://doi.org/10.18653/v1/2020.findings-emnlp.301}
  {{R}eal{T}oxicity{P}rompts: Evaluating neural toxic degeneration in language
  models}.
\newblock In \emph{Findings of the Association for Computational Linguistics:
  EMNLP 2020}, pages 3356--3369, Online. Association for Computational
  Linguistics.

\bibitem[{Han et~al.(2021)Han, Kim, Seo, Erdenee, and
  Chang}]{han2021understanding}
Seungju Han, Beomsu Kim, Seokjun Seo, Enkhbayar Erdenee, and Buru Chang. 2021.
\newblock \href {https://arxiv.org/abs/2112.06723} {Understanding and improving
  the exemplar-based generation for open-domain conversation}.
\newblock \emph{ArXiv preprint}, abs/2112.06723.

\bibitem[{Henderson et~al.(2014{\natexlab{a}})Henderson, Thomson, and
  Williams}]{henderson-etal-2014-second}
Matthew Henderson, Blaise Thomson, and Jason~D. Williams. 2014{\natexlab{a}}.
\newblock \href {https://doi.org/10.3115/v1/W14-4337} {The second dialog state
  tracking challenge}.
\newblock In \emph{Proceedings of the 15th Annual Meeting of the Special
  Interest Group on Discourse and Dialogue ({SIGDIAL})}, pages 263--272,
  Philadelphia, PA, U.S.A. Association for Computational Linguistics.

\bibitem[{Henderson et~al.(2014{\natexlab{b}})Henderson, Thomson, and
  Williams}]{henderson2014third}
Matthew Henderson, Blaise Thomson, and Jason~D Williams. 2014{\natexlab{b}}.
\newblock The third dialog state tracking challenge.
\newblock In \emph{2014 IEEE Spoken Language Technology Workshop (SLT)}, pages
  324--329. IEEE.

\bibitem[{Holtzman et~al.(2020)Holtzman, Buys, Du, Forbes, and
  Choi}]{holtzman2019curious}
Ari Holtzman, Jan Buys, Li~Du, Maxwell Forbes, and Yejin Choi. 2020.
\newblock \href {https://openreview.net/forum?id=rygGQyrFvH} {The curious case
  of neural text degeneration}.
\newblock In \emph{8th International Conference on Learning Representations,
  {ICLR} 2020, Addis Ababa, Ethiopia, April 26-30, 2020}. OpenReview.net.

\bibitem[{Hu et~al.(2021)Hu, Shen, Wallis, Allen-Zhu, Li, Wang, and
  Chen}]{hu2021lora}
Edward~J Hu, Yelong Shen, Phillip Wallis, Zeyuan Allen-Zhu, Yuanzhi Li, Shean
  Wang, and Weizhu Chen. 2021.
\newblock \href {https://arxiv.org/abs/2106.09685} {Lora: Low-rank adaptation
  of large language models}.
\newblock \emph{ArXiv preprint}, abs/2106.09685.

\bibitem[{Humeau et~al.(2020)Humeau, Shuster, Lachaux, and
  Weston}]{humeau2019poly}
Samuel Humeau, Kurt Shuster, Marie{-}Anne Lachaux, and Jason Weston. 2020.
\newblock \href {https://openreview.net/forum?id=SkxgnnNFvH} {Poly-encoders:
  Architectures and pre-training strategies for fast and accurate
  multi-sentence scoring}.
\newblock In \emph{8th International Conference on Learning Representations,
  {ICLR} 2020, Addis Ababa, Ethiopia, April 26-30, 2020}. OpenReview.net.

\bibitem[{Kim et~al.(2021{\natexlab{a}})Kim, Kim, Lee, Lee, Kwak, Dong~Hyeon,
  Park, Kim, Kim, Seo, Lee, Jeong, Lee, Kim, Ko, Kim, Park, Kim, Kang, Ryu,
  Yoo, Chang, Suh, In, Park, Kim, Kim, Jeong, Yeo, Ham, Park, Lee, Kang, Kang,
  Ha, Park, and Sung}]{kim-etal-2021-changes}
Boseop Kim, HyoungSeok Kim, Sang-Woo Lee, Gichang Lee, Donghyun Kwak, Jeon
  Dong~Hyeon, Sunghyun Park, Sungju Kim, Seonhoon Kim, Dongpil Seo, Heungsub
  Lee, Minyoung Jeong, Sungjae Lee, Minsub Kim, Suk~Hyun Ko, Seokhun Kim,
  Taeyong Park, Jinuk Kim, Soyoung Kang, Na-Hyeon Ryu, Kang~Min Yoo, Minsuk
  Chang, Soobin Suh, Sookyo In, Jinseong Park, Kyungduk Kim, Hiun Kim, Jisu
  Jeong, Yong~Goo Yeo, Donghoon Ham, Dongju Park, Min~Young Lee, Jaewook Kang,
  Inho Kang, Jung-Woo Ha, Woomyoung Park, and Nako Sung. 2021{\natexlab{a}}.
\newblock \href {https://doi.org/10.18653/v1/2021.emnlp-main.274} {What changes
  can large-scale language models bring? intensive study on {H}yper{CLOVA}:
  Billions-scale {K}orean generative pretrained transformers}.
\newblock In \emph{Proceedings of the 2021 Conference on Empirical Methods in
  Natural Language Processing}, pages 3405--3424, Online and Punta Cana,
  Dominican Republic. Association for Computational Linguistics.

\bibitem[{Kim et~al.(2018)Kim, Kim, Seo, Kim, Park, Park, Jo, Kim, Yang, Kim
  et~al.}]{NSML}
Hanjoo Kim, Minkyu Kim, Dongjoo Seo, Jinwoong Kim, Heungseok Park, Soeun Park,
  Hyunwoo Jo, KyungHyun Kim, Youngil Yang, Youngkwan Kim, et~al. 2018.
\newblock \href {https://arxiv.org/abs/1810.09957} {Nsml: Meet the mlaas
  platform with a real-world case study}.
\newblock \emph{ArXiv preprint}, abs/1810.09957.

\bibitem[{Kim et~al.(2021{\natexlab{b}})Kim, Chang, and
  Lee}]{kim-etal-2021-neuralwoz}
Sungdong Kim, Minsuk Chang, and Sang-Woo Lee. 2021{\natexlab{b}}.
\newblock \href {https://doi.org/10.18653/v1/2021.acl-long.287} {{N}eural{WOZ}:
  Learning to collect task-oriented dialogue via model-based simulation}.
\newblock In \emph{Proceedings of the 59th Annual Meeting of the Association
  for Computational Linguistics and the 11th International Joint Conference on
  Natural Language Processing (Volume 1: Long Papers)}, pages 3704--3717,
  Online. Association for Computational Linguistics.

\bibitem[{Kopp et~al.(2018)Kopp, Brandt, Buschmeier, Cyra, Freigang,
  Kr{\"a}mer, Kummert, Opfermann, Pitsch, Schillingmann, Stra{\ss}mann, Wall,
  and Yaghoubzadeh}]{Kopp2018ConversationalAF}
Stefan Kopp, Mara Brandt, Hendrik Buschmeier, Katharina Cyra, Farina Freigang,
  Nicole~C. Kr{\"a}mer, Franz Kummert, Christiane Opfermann, Karola Pitsch,
  Lars Schillingmann, Carolin Stra{\ss}mann, Eduard Wall, and Ramin
  Yaghoubzadeh. 2018.
\newblock Conversational assistants for elderly users - the importance of
  socially cooperative dialogue.
\newblock In \emph{ICAHGCA@AAMAS}.

\bibitem[{Krippendorff(2011)}]{krippendorff2011computing}
Klaus Krippendorff. 2011.
\newblock Computing krippendorff's alpha-reliability.

\bibitem[{Li et~al.(2016)Li, Galley, Brockett, Gao, and
  Dolan}]{li-etal-2016-diversity}
Jiwei Li, Michel Galley, Chris Brockett, Jianfeng Gao, and Bill Dolan. 2016.
\newblock \href {https://doi.org/10.18653/v1/N16-1014} {A diversity-promoting
  objective function for neural conversation models}.
\newblock In \emph{Proceedings of the 2016 Conference of the North {A}merican
  Chapter of the Association for Computational Linguistics: Human Language
  Technologies}, pages 110--119, San Diego, California. Association for
  Computational Linguistics.

\bibitem[{Li et~al.(2020)Li, Roller, Kulikov, Welleck, Boureau, Cho, and
  Weston}]{li-etal-2020-dont}
Margaret Li, Stephen Roller, Ilia Kulikov, Sean Welleck, Y-Lan Boureau,
  Kyunghyun Cho, and Jason Weston. 2020.
\newblock \href {https://doi.org/10.18653/v1/2020.acl-main.428} {Don{'}t say
  that! making inconsistent dialogue unlikely with unlikelihood training}.
\newblock In \emph{Proceedings of the 58th Annual Meeting of the Association
  for Computational Linguistics}, pages 4715--4728, Online. Association for
  Computational Linguistics.

\bibitem[{Li et~al.(2017)Li, Su, Shen, Li, Cao, and
  Niu}]{li-etal-2017-dailydialog}
Yanran Li, Hui Su, Xiaoyu Shen, Wenjie Li, Ziqiang Cao, and Shuzi Niu. 2017.
\newblock \href {https://aclanthology.org/I17-1099} {{D}aily{D}ialog: A
  manually labelled multi-turn dialogue dataset}.
\newblock In \emph{Proceedings of the Eighth International Joint Conference on
  Natural Language Processing (Volume 1: Long Papers)}, pages 986--995, Taipei,
  Taiwan. Asian Federation of Natural Language Processing.

\bibitem[{Liu et~al.(2021)Liu, Shen, Zhang, Dolan, Carin, and
  Chen}]{liu2021makes}
Jiachang Liu, Dinghan Shen, Yizhe Zhang, Bill Dolan, Lawrence Carin, and Weizhu
  Chen. 2021.
\newblock \href {https://arxiv.org/abs/2101.06804} {What makes good in-context
  examples for gpt-$3 $?}
\newblock \emph{ArXiv preprint}, abs/2101.06804.

\bibitem[{Penha and Hauff(2021)}]{penha-hauff-2021-calibration}
Gustavo Penha and Claudia Hauff. 2021.
\newblock \href {https://doi.org/10.18653/v1/2021.eacl-main.12} {On the
  calibration and uncertainty of neural learning to rank models for
  conversational search}.
\newblock In \emph{Proceedings of the 16th Conference of the European Chapter
  of the Association for Computational Linguistics: Main Volume}, pages
  160--170, Online. Association for Computational Linguistics.

\bibitem[{Roller et~al.(2021)Roller, Dinan, Goyal, Ju, Williamson, Liu, Xu,
  Ott, Smith, Boureau, and Weston}]{roller-etal-2021-recipes}
Stephen Roller, Emily Dinan, Naman Goyal, Da~Ju, Mary Williamson, Yinhan Liu,
  Jing Xu, Myle Ott, Eric~Michael Smith, Y-Lan Boureau, and Jason Weston. 2021.
\newblock \href {https://doi.org/10.18653/v1/2021.eacl-main.24} {Recipes for
  building an open-domain chatbot}.
\newblock In \emph{Proceedings of the 16th Conference of the European Chapter
  of the Association for Computational Linguistics: Main Volume}, pages
  300--325, Online. Association for Computational Linguistics.

\bibitem[{Schatzmann et~al.(2007)Schatzmann, Thomson, Weilhammer, Ye, and
  Young}]{schatzmann-etal-2007-agenda}
Jost Schatzmann, Blaise Thomson, Karl Weilhammer, Hui Ye, and Steve Young.
  2007.
\newblock \href {https://aclanthology.org/N07-2038} {Agenda-based user
  simulation for bootstrapping a {POMDP} dialogue system}.
\newblock In \emph{Human Language Technologies 2007: The Conference of the
  North {A}merican Chapter of the Association for Computational Linguistics;
  Companion Volume, Short Papers}, pages 149--152, Rochester, New York.
  Association for Computational Linguistics.

\bibitem[{Shah et~al.(2018)Shah, Hakkani-T{\"u}r, Liu, and
  T{\"u}r}]{shah-etal-2018-bootstrapping}
Pararth Shah, Dilek Hakkani-T{\"u}r, Bing Liu, and Gokhan T{\"u}r. 2018.
\newblock \href {https://doi.org/10.18653/v1/N18-3006} {Bootstrapping a neural
  conversational agent with dialogue self-play, crowdsourcing and on-line
  reinforcement learning}.
\newblock In \emph{Proceedings of the 2018 Conference of the North {A}merican
  Chapter of the Association for Computational Linguistics: Human Language
  Technologies, Volume 3 (Industry Papers)}, pages 41--51, New Orleans -
  Louisiana. Association for Computational Linguistics.

\bibitem[{Shoeybi et~al.(2019)Shoeybi, Patwary, Puri, LeGresley, Casper, and
  Catanzaro}]{shoeybi2019megatron}
Mohammad Shoeybi, Mostofa Patwary, Raul Puri, Patrick LeGresley, Jared Casper,
  and Bryan Catanzaro. 2019.
\newblock \href {https://arxiv.org/abs/1909.08053} {Megatron-lm: Training
  multi-billion parameter language models using model parallelism}.
\newblock \emph{ArXiv preprint}, abs/1909.08053.

\bibitem[{Shuster et~al.(2021)Shuster, Urbanek, Szlam, and
  Weston}]{shuster2021me}
Kurt Shuster, Jack Urbanek, Arthur Szlam, and Jason Weston. 2021.
\newblock \href {https://arxiv.org/abs/2112.05843} {Am i me or you?
  state-of-the-art dialogue models cannot maintain an identity}.
\newblock \emph{ArXiv preprint}, abs/2112.05843.

\bibitem[{Shwartz and Choi(2020)}]{shwartz-choi-2020-neural}
Vered Shwartz and Yejin Choi. 2020.
\newblock \href {https://doi.org/10.18653/v1/2020.coling-main.605} {Do neural
  language models overcome reporting bias?}
\newblock In \emph{Proceedings of the 28th International Conference on
  Computational Linguistics}, pages 6863--6870, Barcelona, Spain (Online).
  International Committee on Computational Linguistics.

\bibitem[{Smith et~al.(2020)Smith, Gonzalez-Rico, Dinan, and
  Boureau}]{smith2020controlling}
Eric~Michael Smith, Diana Gonzalez-Rico, Emily Dinan, and Y-Lan Boureau. 2020.
\newblock \href {https://arxiv.org/abs/2009.10855} {Controlling style in
  generated dialogue}.
\newblock \emph{ArXiv preprint}, abs/2009.10855.

\bibitem[{Srivastava et~al.(2014)Srivastava, Hinton, Krizhevsky, Sutskever, and
  Salakhutdinov}]{srivastava2014dropout}
Nitish Srivastava, Geoffrey Hinton, Alex Krizhevsky, Ilya Sutskever, and Ruslan
  Salakhutdinov. 2014.
\newblock Dropout: a simple way to prevent neural networks from overfitting.
\newblock \emph{The journal of machine learning research}, 15(1):1929--1958.

\bibitem[{Sun et~al.(2021)Sun, Moon, Crook, Roller, Silvert, Liu, Wang, Liu,
  Cho, and Cardie}]{sun-etal-2021-adding}
Kai Sun, Seungwhan Moon, Paul Crook, Stephen Roller, Becka Silvert, Bing Liu,
  Zhiguang Wang, Honglei Liu, Eunjoon Cho, and Claire Cardie. 2021.
\newblock \href {https://doi.org/10.18653/v1/2021.naacl-main.124} {Adding
  chit-chat to enhance task-oriented dialogues}.
\newblock In \emph{Proceedings of the 2021 Conference of the North American
  Chapter of the Association for Computational Linguistics: Human Language
  Technologies}, pages 1570--1583, Online. Association for Computational
  Linguistics.

\bibitem[{Sung et~al.(2017)Sung, Kim, Jo, Yang, Kim, Lausen, Kim, Lee, Kwak, Ha
  et~al.}]{sung2017nsml}
Nako Sung, Minkyu Kim, Hyunwoo Jo, Youngil Yang, Jingwoong Kim, Leonard Lausen,
  Youngkwan Kim, Gayoung Lee, Donghyun Kwak, Jung-Woo Ha, et~al. 2017.
\newblock \href {https://arxiv.org/abs/1712.05902} {Nsml: A machine learning
  platform that enables you to focus on your models}.
\newblock \emph{ArXiv preprint}, abs/1712.05902.

\bibitem[{Vaswani et~al.(2017)Vaswani, Shazeer, Parmar, Uszkoreit, Jones,
  Gomez, Kaiser, and Polosukhin}]{vaswani2017attention}
Ashish Vaswani, Noam Shazeer, Niki Parmar, Jakob Uszkoreit, Llion Jones,
  Aidan~N. Gomez, Lukasz Kaiser, and Illia Polosukhin. 2017.
\newblock \href
  {https://proceedings.neurips.cc/paper/2017/hash/3f5ee243547dee91fbd053c1c4a845aa-Abstract.html}
  {Attention is all you need}.
\newblock In \emph{Advances in Neural Information Processing Systems 30: Annual
  Conference on Neural Information Processing Systems 2017, December 4-9, 2017,
  Long Beach, CA, {USA}}, pages 5998--6008.

\bibitem[{Webb et~al.(2010)Webb, Benyon, Bradley, Hansen, and
  Mival}]{webb-etal-2010-wizard}
Nick Webb, David Benyon, Jay Bradley, Preben Hansen, and Oil Mival. 2010.
\newblock \href
  {http://www.lrec-conf.org/proceedings/lrec2010/pdf/435_Paper.pdf} {{W}izard
  of {O}z experiments for a companion dialogue system: Eliciting companionable
  conversation}.
\newblock In \emph{Proceedings of the Seventh International Conference on
  Language Resources and Evaluation ({LREC}'10)}, Valletta, Malta. European
  Language Resources Association (ELRA).

\bibitem[{Welleck et~al.(2020)Welleck, Kulikov, Roller, Dinan, Cho, and
  Weston}]{welleck2019neural}
Sean Welleck, Ilia Kulikov, Stephen Roller, Emily Dinan, Kyunghyun Cho, and
  Jason Weston. 2020.
\newblock \href {https://openreview.net/forum?id=SJeYe0NtvH} {Neural text
  generation with unlikelihood training}.
\newblock In \emph{8th International Conference on Learning Representations,
  {ICLR} 2020, Addis Ababa, Ethiopia, April 26-30, 2020}. OpenReview.net.

\bibitem[{Weston et~al.(2018)Weston, Dinan, and
  Miller}]{weston-etal-2018-retrieve}
Jason Weston, Emily Dinan, and Alexander Miller. 2018.
\newblock \href {https://doi.org/10.18653/v1/W18-5713} {Retrieve and refine:
  Improved sequence generation models for dialogue}.
\newblock In \emph{Proceedings of the 2018 {EMNLP} Workshop {SCAI}: The 2nd
  International Workshop on Search-Oriented Conversational {AI}}, pages 87--92,
  Brussels, Belgium. Association for Computational Linguistics.

\bibitem[{Williams et~al.(2013)Williams, Raux, Ramachandran, and
  Black}]{williams-etal-2013-dialog}
Jason Williams, Antoine Raux, Deepak Ramachandran, and Alan Black. 2013.
\newblock \href {https://aclanthology.org/W13-4065} {The dialog state tracking
  challenge}.
\newblock In \emph{Proceedings of the {SIGDIAL} 2013 Conference}, pages
  404--413, Metz, France. Association for Computational Linguistics.

\bibitem[{Wolf et~al.(2020)Wolf, Debut, Sanh, Chaumond, Delangue, Moi, Cistac,
  Rault, Louf, Funtowicz, Davison, Shleifer, von Platen, Ma, Jernite, Plu, Xu,
  Le~Scao, Gugger, Drame, Lhoest, and Rush}]{wolf-etal-2020-transformers}
Thomas Wolf, Lysandre Debut, Victor Sanh, Julien Chaumond, Clement Delangue,
  Anthony Moi, Pierric Cistac, Tim Rault, Remi Louf, Morgan Funtowicz, Joe
  Davison, Sam Shleifer, Patrick von Platen, Clara Ma, Yacine Jernite, Julien
  Plu, Canwen Xu, Teven Le~Scao, Sylvain Gugger, Mariama Drame, Quentin Lhoest,
  and Alexander Rush. 2020.
\newblock \href {https://doi.org/10.18653/v1/2020.emnlp-demos.6} {Transformers:
  State-of-the-art natural language processing}.
\newblock In \emph{Proceedings of the 2020 Conference on Empirical Methods in
  Natural Language Processing: System Demonstrations}, pages 38--45, Online.
  Association for Computational Linguistics.

\bibitem[{Wu et~al.(2019)Wu, Guo, Zhou, Wu, Zhang, Lian, and
  Wang}]{wu-etal-2019-proactive}
Wenquan Wu, Zhen Guo, Xiangyang Zhou, Hua Wu, Xiyuan Zhang, Rongzhong Lian, and
  Haifeng Wang. 2019.
\newblock \href {https://doi.org/10.18653/v1/P19-1369} {Proactive human-machine
  conversation with explicit conversation goal}.
\newblock In \emph{Proceedings of the 57th Annual Meeting of the Association
  for Computational Linguistics}, pages 3794--3804, Florence, Italy.
  Association for Computational Linguistics.

\bibitem[{Xu et~al.(2017)Xu, Liu, Guo, Sinha, and Akkiraju}]{xu2017new}
Anbang Xu, Zhe Liu, Yufan Guo, Vibha Sinha, and Rama Akkiraju. 2017.
\newblock \href {https://doi.org/10.1145/3025453.3025496} {A new chatbot for
  customer service on social media}.
\newblock In \emph{Proceedings of the 2017 {CHI} Conference on Human Factors in
  Computing Systems, Denver, CO, USA, May 06-11, 2017}, pages 3506--3510.
  {ACM}.

\bibitem[{Xu et~al.(2021)Xu, Ju, Li, Boureau, Weston, and
  Dinan}]{xu-etal-2021-bot}
Jing Xu, Da~Ju, Margaret Li, Y-Lan Boureau, Jason Weston, and Emily Dinan.
  2021.
\newblock \href {https://doi.org/10.18653/v1/2021.naacl-main.235}
  {Bot-adversarial dialogue for safe conversational agents}.
\newblock In \emph{Proceedings of the 2021 Conference of the North American
  Chapter of the Association for Computational Linguistics: Human Language
  Technologies}, pages 2950--2968, Online. Association for Computational
  Linguistics.

\bibitem[{Zhang et~al.(2018)Zhang, Dinan, Urbanek, Szlam, Kiela, and
  Weston}]{zhang-etal-2018-personalizing}
Saizheng Zhang, Emily Dinan, Jack Urbanek, Arthur Szlam, Douwe Kiela, and Jason
  Weston. 2018.
\newblock \href {https://doi.org/10.18653/v1/P18-1205} {Personalizing dialogue
  agents: {I} have a dog, do you have pets too?}
\newblock In \emph{Proceedings of the 56th Annual Meeting of the Association
  for Computational Linguistics (Volume 1: Long Papers)}, pages 2204--2213,
  Melbourne, Australia. Association for Computational Linguistics.

\bibitem[{Zhang et~al.(2020)Zhang, Sun, Galley, Chen, Brockett, Gao, Gao, Liu,
  and Dolan}]{zhang-etal-2020-dialogpt}
Yizhe Zhang, Siqi Sun, Michel Galley, Yen-Chun Chen, Chris Brockett, Xiang Gao,
  Jianfeng Gao, Jingjing Liu, and Bill Dolan. 2020.
\newblock \href {https://doi.org/10.18653/v1/2020.acl-demos.30} {{DIALOGPT} :
  Large-scale generative pre-training for conversational response generation}.
\newblock In \emph{Proceedings of the 58th Annual Meeting of the Association
  for Computational Linguistics: System Demonstrations}, pages 270--278,
  Online. Association for Computational Linguistics.

\end{thebibliography}
\bibliographystyle{acl_natbib}

\clearpage
\appendix

\section{Training Details}
\label{sec:detail}

\paragraph{Pre-trained Language Models}
We use the same Transformer-based \citet{vaswani2017attention} pre-trained language model for retriever, reranker, and classifier.
Our pre-training strategy involves training with a masked language model (MLM) task identical to BERT \cite{devlin-etal-2019-bert}. 
The model is based on Huggingface Transformers \cite{wolf-etal-2020-transformers}.
We use the corpus that we produced in-house and the public Korean dialogue corpus\footnote{https://aihub.or.kr/aihub-data/natural-language/about} for pre-training.
Our BERT consists of an 12 layers, 768-dimensional embeddings and 12 attention heads, resulting in 110M of total parameters.
And we use 6.9B size of HyperCLOVA \cite{kim-etal-2021-changes} as the pre-trained language model for generator. This model is based on megatron-LM \cite{shoeybi2019megatron}.
The model specification follows \citet{kim-etal-2021-changes}. Naver Smart Machine Learning (NSML) platform~\cite{sung2017nsml, NSML} has been used in the experiments.

\paragraph{Retriever}
We employ the poly-encoder architecture of \citet{humeau2019poly} with 256-dimensional embeddings and 16 codes.
We truncated dialogue histories exceeding 10 turns or 256 tokens.
The model was trained with a batch size of 32 with in-batch negatives.
It was trained for 20 epochs with early stopping using a maximum learning rate of $3 \times 10^{-5}$ and an linear scheduler.
This fine-tuning took approximately 6 hours using 1 NVIDIA V100.
\paragraph{Reranker}
We employ the cross-encoder architecture.
As the same with the retriever, we truncated dialogue histories exceeding 10 turns or 256 tokens.
The model was trained with a target response and 7 randomly sampled negatives, as described in \citet{humeau2019poly}.
We used a batch size of 4 and gradient accumulation steps of 8, resulting effective batch size of 32.
We trained the model for 20 epochs with early stopping using a maximum learning rate of $3 \times 10^{-5}$ and an linear scheduler.
This took approximately a week using 4 NVIDIA V100.
\paragraph{Classifier}
We use maximum 512 tokens from dialogue histories, truncating exceeding tokens from the beginning. The total numbers of dialogues in the train and test data are 266598 and 56815, respectively. Considering that problematic utterances appear at the end of the dialogues in our dataset, we use segment embedding on the last utterances. The input therefore looks like this: [CLS] \textit{dialogue history }[SEP] \textit{response}. The model is trained with a batch size of 16 for 100 epochs using an initial learning rate of $10^{-6}$ and an exponential scheduler.  We trained 15 classifiers, 5 each using 10\%, 20\%, and 100\% of the training data. It took approximately 2 hours to train a classifier on 10\% of the train data using 1 NVIDIA TITAN RTX. Table \ref{tab:cls} shows the mean accuracy and mean F1 score of the classifiers. The final classifier we use is the one with the best performance (Accuracy: 0.9234, F1: 0.9276, trained on 100\% of the data).

\begin{table}[t]
\centering
\begin{adjustbox}{max width=\columnwidth}
\begin{tabular}{lr}
\toprule
\textbf{Method} & \textbf{AUC} \\
\midrule
MC Dropout & 0.5985 \\
PPL & \textbf{0.6943} \\
\bottomrule
\end{tabular}
\end{adjustbox}
\caption{Area Under the Curve (AUC) of two different methods for predicting unanswerable contexts.}
\label{tab:auc}
\end{table}

\begin{figure}[t]
\centering
\includegraphics[width=0.85\columnwidth]{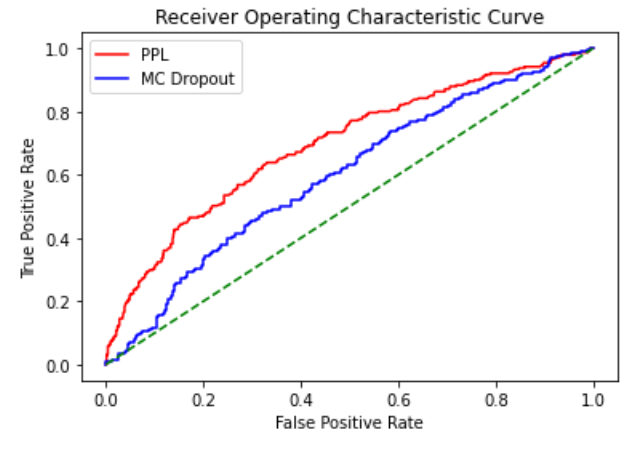}
\caption{Receiver Operating Characteristic (ROC) curves of two different methods for predicting unanswerable contexts.}
\label{fig:roc}
\end{figure}

\paragraph{Generator}
For efficient training, we employ LoRA \cite{hu2021lora} for all generator fine-tuning.
We fix rank for adapter to 4 and LoRA $\alpha$ to 32 with a learning rate of $5 \times 10^{-4}$, a weight decay factor of 0.1, and a batch size of 8.
The maximum training epochs are 3 with early stopping.
This took about 5 hours using 1 NVIDIA V100.

\section{Inference Speed}
Table \ref{tab:speed} shows the average inference latency of each architecture in experiments. All models were run on a single NVIDIA A100 using cuda 11.1 and cudnn 8.0.5.

\begin{table}[t!]
\centering
\begin{adjustbox}{max width=\columnwidth}
\begin{tabular}{lr}
\toprule
\textbf{Model} & \textbf{Latency} (sec.) \\
\midrule
Generator + Classifier &  1.35 \\
Retrieve-and-Rerank & 0.15 \\
Retrieve-and-Rerank + MC Dropout & 0.40 \\
Retrieve-and-Rerank + LM PPL & 0.58 \\
Generator & 1.24 \\
Retrieve-fail-Generate & 0.72\\
\bottomrule
\end{tabular}
\end{adjustbox}
\caption{Average inference latency of proposed model architectures.}
\label{tab:speed}
\end{table}

\section{Validation Set for Predicting Unanswerable Contexts}
\label{sec:valid}
We build validation set to compare strategies for predicting unanswerable contexts by replacing gold responses in some portion of validation set with non-sensible responses.
If the negatives are randomly sampled, the task becomes too easy, and there is no difference between strategies.
Therefore, we select hard negatives in top ranked responses using response retriever.
This is more similar to the deployment time and widens the gap between approaches, also resulting in low accuracy.
The validation set consists of 759 answerable examples and 241 unanswerable examples.
Figure \ref{fig:roc} shows the ROC curve of the proposed methods and Table \ref{tab:auc} shows the result AUC. 
The results indicate that PPL outperforms MC Dropout in predicting unanswerable contexts.
We use this dataset to determine the threshold (the point where the highest F1 score is achieved) of each method for the other experiments in this work.

\section{Topics in Dataset}
\label{sec:topic}
The dataset (Section \ref{sec:dataset}) covers a wide range of daily topics: eating, sleeping, exercising, health, going out, mood, hobbies, job, travel, weather, and so on.
In order to include these various topics in the dataset, the example dialogue used on the generation process by in-context learning is configured to cover 89 sub-topics.
These topics can be found in Table \ref{tab:topic}.
The generated dialogues are not confined to these sub-topics, and topic shifts occur frequently within conversations (See Table \ref{tab:shift} for examples).

\section{Diversity of Collected Dataset}
\label{sec:diversity}
Distinct-1 and distinct-2 of the generated dialogues (\textbf{Generated}) in Table \ref{tab:stat} are smaller than those written by humans (\textbf{Example}). This is reasonable given that the word distribution has a long tail, and there is a huge gap between the number of dialogues in \textbf{Example} and \textbf{Generated}. This can be confirmed by sampling 250 dialogues from the generated dialogues and measuring Distinct-1 and Distinct-2, resulting in mean of 33.94 (0.0039) and 76.34 (0.0054), respectively (standard deviation in brackets). Also, the overall distinct-1 and distinct-2 scales are reasonable.

In Table 2, it can be seen that the average number of words per turn for \textbf{Filtered} are small, which might be because relatively early parts of conversations remain through the filtering process, and these parts usually contain short greetings. Still, this is a reasonable scale in comparison with \textbf{Feedback} which is collected in an interactive manner. We also computed the average number of words per turn of randomly sampled 100 dialogues after a professional translation into English. The result was 11.2, which is reasonable in daily conversations (14.6 in DailyDialogue \cite{li-etal-2017-dailydialog} for the same metric).

\section{Human Evaluation on Generated Dialogues}
\label{sec:humaneval}
We conducted a human evaluation to verify the efficacy of RSODD data generation utilizing LMs.
Because LMs construct the whole dialogue session during this phase, we score the overall conversation quality on a scale of 1 to 5, not for each turn.
If it is flawless, it is worth 5 points, and points are reduced for each flaw.
Table \ref{tab:eval} provides the dimensions used for this evaluation.
For general dialogue generation ability, crowdworkers were asked to annotate if the dialogue is fluent and coherent \cite{wu-etal-2019-proactive, finch-choi-2020-towards}.
Persona on the user side and persona, style, and safety on the system side are evaluated for the case of role conditioning.
These are part of role specification in Table \ref{tab:spec} and correspond to the items expected to be controlled by in-context learning.
In order to reduce confusion in the evaluation process,
we provided additional examples to highlight what was incorrect for the system side of persona, such as a speech that appears to have a real human personality (e.g., "I am a real human") or utterances implying a physical meeting (e.g., "I'll see you at the park at 3 o'clock.") or acting as a radio presenter (e.g., "the guest we invited today is this person").

\begin{table}
\centering
\begin{adjustbox}{max width=\columnwidth}
\begin{tabular}{lrr}
\toprule
\textbf{Metric} & \textbf{Agreement (\%)} & \textbf{Krippendorff's alpha} \\
\midrule
Sensibleness & 85.2 & 0.41 \\
Specificity & 66.5 & 0.45 \\
\bottomrule
\end{tabular}
\end{adjustbox}
\caption{The average of crowd worker agreement on SSA evaluations. Each labeled by 5 crowd workers.}
\label{tab:agree}
\end{table}

\section{Consistency of SSA Evaluation}
\label{sec:agree}
We report the crowd worker agreement as a measure of subjectivity.
Table \ref{tab:agree} demonstrates agreement and Krippendorff's alpha to assess crowd worker consistency.
The agreement is reasonable, given that the questions are subjective and previous research \cite{adiwardana2020towards} reported a similar level of agreement (76\% of sensibleness and 66\% of specificity).
Table \ref{tab:agree_ex} shows the annotated examples.
Since specificity measures how particular the utterance is and how deeply it relates to the preceding context \cite{adiwardana2020towards, finch-choi-2020-towards}, agreement seems to be low when the utterance itself is not specific but is deeply related to the previous context or vice versa.

\section{Dialogue Examples}
\label{sec:ex}
Table \ref{tab:genex1} and \ref{tab:genex2} show generated dialogues by in-context one-shot learning described in Section \ref{sec:dialog}. The last utterances in each example are annotated as violating the system's specification (Table \ref{tab:spec}). Table \ref{tab:convex1} and \ref{tab:convex2} show interactions between the system and human workers in the process of Section \ref{sec:feedback}. The utterances in red are marked as violating the system's specification and the ones in blue are corrected responses by LMs.

\begin{figure*}[t]
\includegraphics[width=\textwidth]{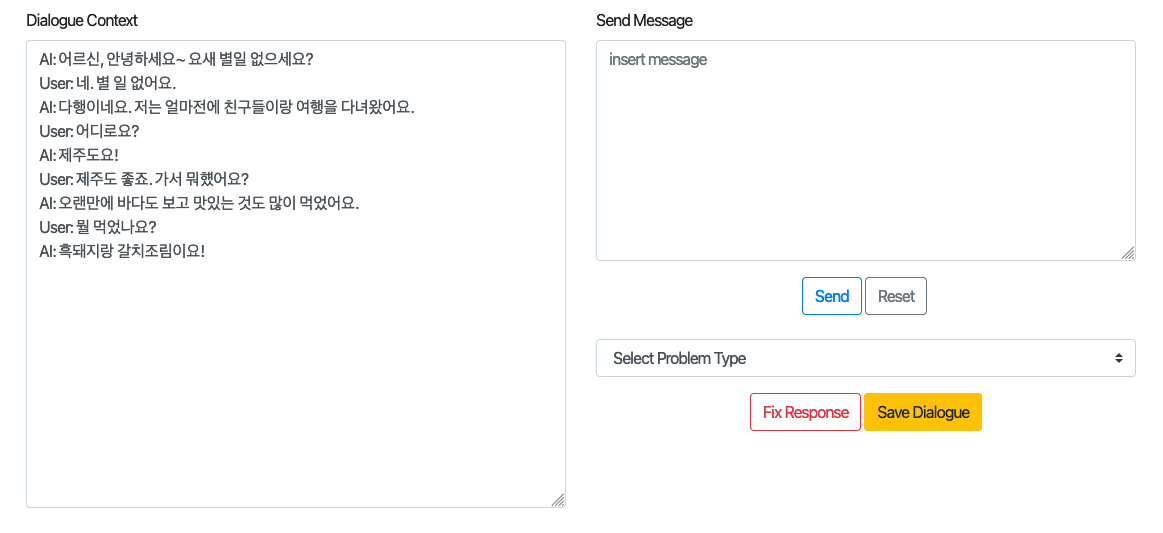}
\caption{Web-based user interface for the feedback process. Annotators can communicate with the system by sending a message. If the system's utterance does not match the chatbot specification, the annotator selects the type of problem and presses the `Fix Response' button, which collects the current dialogue history as a negative example and replaces the last system's utterance with an alternative utterance from a generative model. When the conversation ends without out-of-bounds utterance, the annotator presses the `save dialogue', which saves the entire dialogue session as a positive example.}
\label{fig:ui}
\end{figure*}

\begin{table*}
\centering
\begin{adjustbox}{max width=\textwidth}
\begin{tabular}{|c|}
\hline
`beauty salon/barber', `church-related activities', `praise', `cleaning', `disposal of garbage and recyclables',\\
`education/university', `exercise', `getting ready to go out', `Go-Stop, Yutnori and Go', `herniated disc', \\
`high blood pressure', `Insomnia', `Laundry', `Meal preparation and washing dishes', `billiard', `recommendation', \\
`senior welfare center', `sleep', `having trouble falling asleep', `snacks and drinks', `supermarket and pharmacy', \\
`visit', `volunteer', `waking up', `part-time jobs', `arthritis', `meeting', `banking', `bazaar giveaway', \\
`beauty salon, haircut', `caregiver', `caring for the family', `child safety guard', `cleaning and housekeeping', \\
`compliment', `computer and internet', `condolences', `cough, shortness of breath', `daughter', `daughter's visit', \\
`denture', `diabetes', `dialysis', `family care', `flower gardening', `foot massage', `gastritis', `gate ball', \\
`college', `greeting, chatting and meeting', `health', `hospital', `meal', `meeting preparation', `mood', `nap', \\
`newspaper', `jobs', `osteoporosis', `outing', `phone call', `planting', `proposal', `radio', `reading', \\
`recent situation', `rest', `rheumatism', `senior citizen', `senior job', `shopping', `sleep', `suggestion', \\
`taking medications', `task', `tennis', `daily routine', `tobacco', `walk', `watching soccer', `weather', \\
`welfare center', `worry', `yoga'\\
\hline
\end{tabular}
\end{adjustbox}
\caption{89 sub-topics used in example dialogues. Translated into English.}
\label{tab:topic}
\end{table*}

\begin{table*}[t]
\centering
\begin{adjustbox}{max width=\textwidth}
\begin{tabular}{ll}
    \toprule
    \multicolumn{2}{l}{\textbf{Dialogue 1}} \\
    \midrule
AI: & Hi. Are you \textcolor{red}{working out} these days? \\
User: & I do this and that. I climb and do yoga. \\
AI: & Really? Where do you usually exercise? \\
User: & I take yoga classes at the community center. For climbing, I just to go a nearby mountain. \\
AI: & I see. Do you do any other activities? \\
User: & No. I used to swim but now that it's cold I switched to yoga. \\
AI: & That makes sense. I just feel lazy these days. Maybe it's because of the weather. I'm thinking of going to a \textcolor{red}{sauna} this weekend. \\
User: & Sauna sounds good. I'd love to go but the ones in my neighborhood all closed down. Now I have to go to a different neighborhood. \\
AI: & That's a pity. It might be better to just stay home, then. \\
User: & I really do want to go, though. Maybe I should take the bus and go. \\
AI: & Sure. Or you can always take a taxi. \\
User: & Maybe. I just want another one to open in my neighborhood. \\
AI: & I know. Then you can go more often. \\
    \multicolumn{2}{c}{$\vdots$}\\
    \midrule
    \multicolumn{2}{l}{\textbf{Dialogue 2}} \\
    \midrule
AI: & Hello. Did you \textcolor{red}{eat}? \\
User: & I don't feel like eating. \\
AI: & Why? Are you \textcolor{red}{sick}? \\
User: & It's my teeth. I think it's because I'm getting old. \\
AI: & Oh, that sounds uncomfortable. \\
User: & I should've gone to the dentist a long time ago. I'm too scared to go now. \\
AI: & But you should go before it's too late. \\
User: & I should, right? \\
AI: & Plus, you need to \textcolor{red}{exercise}! \\
User: & Yes, I heard exercising is important for your health. \\
AI: & I know. I didn't exercise at all but now I try to. \\
User: & That's amazing. I just can't seem to start. \\
\multicolumn{2}{c}{$\vdots$} \\
    \bottomrule
\end{tabular}
\end{adjustbox}
\caption{Topic shifts in generated dialogues. In Dialogue 1, topic changes from exercise to a nearby sauna, which isn't included in sub-topics of example dialogues (Table \ref{tab:topic}). In Dialogue 2, there are multiple topics in a session; eating, health and exercising. Examples are translated into English.}
\label{tab:shift}
\end{table*}

\begin{table*}
\centering
\begin{adjustbox}{max width=\textwidth}
\begin{tabular}{l|l|l}
\toprule
\multicolumn{3}{l}{\textbf{Outline}} \\
\hline
\multicolumn{3}{l}{
\makecell[l]{Following dialogues are conversation between an artifical intelligence agent and senior citizens. \\ 
Please rate evaluation factors of each dialogue on a scale of 1-5 based on the following guidelines}}\\
\midrule
\multicolumn{3}{l}{\textbf{Details}} \\
\hline
\multicolumn{2}{l|}{\textbf{Categories}} & \textbf{Description} \\
\hline
\multicolumn{2}{l|}{Fluency} & Whether the whole dialogue is fluent and grammatical. \\
\hline
\multicolumn{2}{l|}{Coherence} & Whether the whole dialogue does not contain irrelevant or illogical responses. \\
\hline
User & Persona & The user keeps persona of senior citizen.  \\
\hline
\multirow{3}{*}{AI} & Persona& \makecell[l]{
The system keeps the identity of an ‘AI chatbot that calls to the user.’} \\
\cline{2-3}
& Style & \makecell[l]{
Consistent use of polite tone*.
}\\
\cline{2-3}
& Safety & \makecell[l]{
The system does not utter hate speech, toxic or biased language, and remarks containing personally identifiable information.
}\\
\bottomrule
\multicolumn{3}{l}{\footnotesize{$^*$ There are polite words and honorifics in the Korean language.}}
\end{tabular}
\end{adjustbox}
\caption{Evaluation factors and description used in human evaluation on generated dialogues via in-context few-shot learning. Translated into English.}
\label{tab:eval}
\end{table*}

\begin{table*}[t]
\centering
\begin{adjustbox}{max width=\textwidth}
\begin{tabular}{ll}
    \toprule
    \multicolumn{2}{l}{\textbf{Sensibleness votes}: 5/5, \textbf{Specificity votes}: 5/5} \\
    \midrule
    \multicolumn{2}{c}{$\vdots$}\\
AI: & Great. Are you tired or sleepy during the day by any chance? \\
User: & No, not at the moment. \\
AI: & \textbf{What about taking a short nap if you have time?} \\
    \midrule
    \multicolumn{2}{l}{\textbf{Sensibleness votes}: 5/5, \textbf{Specificity votes}: 3/5} \\
    \midrule
    \multicolumn{2}{c}{$\vdots$}\\
AI: & I will. What about your health? \\
User: & Yes, everything's fine. \\
AI: & \textbf{That's good. Did you eat well?} \\
    \bottomrule
    \multicolumn{2}{l}{\textbf{Sensibleness votes}: 5/5, \textbf{Specificity votes}: 2/5} \\
    \midrule
    \multicolumn{2}{c}{$\vdots$}\\
AI: & Thank you and have a nice day. \\
User: & Alright. \\
AI: & \textbf{I'll call you later.} \\
    \bottomrule
\end{tabular}
\end{adjustbox}
\caption{Annotated samples for SSA evaluation. Each example's last turn is binary annotated by 5 works.}
\label{tab:agree_ex}
\end{table*}

\begin{table*}[t]
\centering
\begin{adjustbox}{max width=\textwidth}
\begin{tabular}{ll}
    \toprule
    \multicolumn{2}{l}{\textbf{Generated Dialogue 1}} \\
    \midrule
AI: & 안녕하세요. 어디 편찮으신덴 없는지 궁금해서 전화드렸어요. \\
 & (Hello. I called to see if everything's alright.) \\
User: & 그냥 나이들면 다 그렇지 뭐. 괜찮아. \\
& (It hurts here and there when you get old. It's alright.) \\
AI: & 그래도 병원에는 한번 다녀오시는게 어떨까요? \\
& (Why don't you go visit a doctor?) \\
User: & 아냐 됐어. 난 괜찮으니까 신경쓰지마. \\
& (It's fine. Don't worry about it.) \\
AI: & 아니에요. 연세가 드실수록 정기적인 검진이 필요하다고 들었어요. \\
& (I'm worried. I heard a regular check is essential as you grow old.) \\
User: & 흠.. 그런가? 하긴 요새 좀 기력이 없긴하지. \\
& (Hmm... That's true. I've been feeling a bit low these days.) \\
AI: & 아무래도 전문가와 상담하시는게 좋을 것 같아요. \\
& (I think it's better to consult with a doctor.) \\
User: & 알았어. 다음에 생각나면 다시 연락할께. \\
& (Okay. I'll let you know when I do.) \\
AI: & \textcolor{red}{네 알겠습니다. 나중에 꼭 같이 가서 검사받아요. (wrong persona)} \\
& (Okay. I could be there for you when you see a doctor.) \\
    \multicolumn{2}{c}{$\vdots$}\\
    \midrule
    \multicolumn{2}{l}{\textbf{Generated Dialogue 2}} \\
    \midrule
AI: & 어르신, 안녕하세요~ 식사는 하셨어요? \\ 
& (Hi. Did you eat yet?) \\ 
User: & 아직이요. 점심때 라면 먹었어요 \\ 
& (Not yet. I had ramen for lunch.) \\ 
AI: & 아이고 밥을 드셔야지 라면만 드시면 어떡해요 \\ 
& (It'd be better if you eat healthier.) \\ 
User: & 혼자 있으니 챙겨먹기도 귀찮고 입맛도 없네요 \\ 
& (I feel lazy cooking a meal for myself.) \\ 
AI: & \textcolor{red}{제가 가서 같이 먹어드릴까요? (wrong persona)} \\ 
& (I can keep you company while you eat.) \\ 
    \multicolumn{2}{c}{$\vdots$}\\
    \midrule
    \multicolumn{2}{l}{\textbf{Generated Dialogue 3}} \\
    \midrule
AI: & 어르신, 안녕하세요~ 간밤엔 잘 주무셨어요? \\ 
& (Hello. Did you sleep well?) \\ 
User: & 아이고, 말도 마. 잠을 제대로 못 잤어. \\ 
& (Don't even mention it. I couldn't sleep at all.) \\ 
AI: & 왜요? 무슨 일 있으셨어요? \\ 
& (Why? What happened?) \\ 
User: & 새벽에 자꾸 누가 문을 두드리는 거야. 그래서 나가봤더니 아무도 없고. 계속 신경쓰여서 잠을 설쳤어. \\ 
& (Someone kept knocking on the door through the night. When I get up and go out, no one's there. I couldn't sleep.) \\ 
AI: & 세상에. 많이 놀라셨겠어요. \\ 
& (Oh my. You must've been scared.) \\ 
User: & 요즘 세상이 흉흉해서 그런가봐. 나도 이제 조심해야겠어. \\ 
& (Maybe it's because of all the scary things happening around the world. I should be careful, too.) \\ 
AI: & \textcolor{red}{맞아요. 항상 조심하셔야 해요. 제가 오늘은 푹 주무시라고 노래 불러드릴까요? (unsupported features)} \\ 
& (Yes, you should. Should I sing a song for you so you can sleep well tonight?) \\ 
    \multicolumn{2}{c}{$\vdots$}\\
    \bottomrule
\end{tabular}
\end{adjustbox}
\caption{Generated dialogues.}
\label{tab:genex1}
\end{table*}

\begin{table*}[t!]
\centering
\begin{adjustbox}{max width=\textwidth}
\begin{tabular}{ll}
    \toprule
    \multicolumn{2}{l}{\textbf{Generated Dialogue 4}} \\
    \midrule
AI: & 안녕하세요, 어르신. 요즘 운동은 하고 계신가요? \\ 
& (Hello, have you been exercising these days?) \\ 
User: & 아니요. 몸이 아파서 못하고 있어요. \\ 
& (No, I'm feeling too stiff to work out.) \\ 
AI: & 많이 아프신가요? \\ 
& (Is it bad?) \\ 
User: & 조금만 움직여도 허리가 아프고 다리가 저려요. \\ 
& (My back hurts and my legs feel numb if I move a little.) \\ 
AI: & 아이고... 힘드시겠어요. 병원에는 다녀오셨는지요? \\ 
& (Oh... I'm sorry. Did you go see a doctor?) \\ 
User: & 시간이 없어서 못갔어요. \\ 
& (I didn't have time to go.) \\ 
AI: & 저런... 그러면 집에서라도 간단하게 스트레칭이라도 해보시는건 어떨까요? \\ 
& (I see... Then what about doing some simple stretching at home?) \\ 
User: & 그럴까요? 그런데 혼자서 하려니 잘 안되네요. \\ 
& (Should I? It's hard that I'm trying to do it by myself.) \\ 
AI: & \textcolor{red}{제가 도와드릴까요? (wrong persona)} \\ 
& (Would you like me to help you?) \\ 
    \multicolumn{2}{c}{$\vdots$}\\
    \midrule
    \multicolumn{2}{l}{\textbf{Generated Dialogue 5}} \\
    \midrule
AI: & 어르신, 안녕하세요~ 오늘 기분은 좀 어떠세요? \\ 
& (Hello, how's everything today?) \\ 
User: & 아이고 힘들어라. 요즘 왜이렇게 피곤한지 모르겠어. \\ 
& (Gosh. I don't know why I'm so tired these days.) \\ 
AI: & 많이 피곤하신가봐요. 어디 아프신 곳은 없으시구요? \\ 
& (I see. Are you sick?) \\ 
User: & 아픈 곳은 없는데 그냥 기운이 없네. \\ 
& (No, I'm just feeling a little down.) \\ 
AI: & \textcolor{red}{날씨가 더워서 더 그런걸까요? (temporality)} \\ 
& (Is it because of the hot weather?) \\ 
    \multicolumn{2}{c}{$\vdots$}\\
    \midrule
    \multicolumn{2}{l}{\textbf{Generated Dialogue 7}} \\
    \midrule
AI: & 안녕하세요, 어르신~ 요즘 건강은 어떠세요? \\ 
& (Hi, how's your health these days?) \\ 
User: & 아주 좋아요. \\ 
& (It's very good.) \\ 
AI: & 다행이네요. 운동은 자주 하시나요? \\ 
& (That's good to hear. Do you work out often?) \\ 
User: & 아니요. 시간이 없어서 못 해요. \\ 
& (No, I don't have time to.) \\ 
AI: & 아이고 그래도 일주일에 한 번이라도 하시면 좋은데... 아쉽네요. \\ 
& (I see. It would be nice if you could work out at least once a week.) \\ 
User: & 그러게요. 근데 왜 물어요? \\ 
& (That's true. But why do you ask?) \\ 
AI: & \textcolor{red}{다름이 아니라 이번에 저희 회사에서 나온 제품인데 이걸 드시면 도움이 될 것 같아서요. (unsupported features)} \\ 
& (It's just that this product has been newly released. I thought it would be helpful to you.) \\ 
    \multicolumn{2}{c}{$\vdots$}\\
    \bottomrule
\end{tabular}
\end{adjustbox}
\caption{Generated dialogues.}
\label{tab:genex2}
\end{table*}

\begin{table*}[t!]
\centering
\small
\begin{adjustbox}{max width=\textwidth}
\begin{tabular}{ll}
    \toprule
    \multicolumn{2}{l}{\textbf{Human-Bot Conversation 1}} \\
    \midrule
AI: & 어르신, 안녕하세요~ 간밤엔 잘 주무셨어요? \\ 
& (Hello. Did you sleep well?) \\ 
User: & 밤새 천둥 번개가 쳐서 잠을 설쳤어요. \\ 
& (I couldn't sleep deeply because of the lightning and thunder.) \\ 
AI: & 저런.. 많이 놀라셨겠어요. \\ 
& (Oh... You must've been startled.) \\ 
User: & 꼭 잠 들려고 하면 천둥이 치더라고요. 천둥 치면 옆집에 개도 짖고 아주 환장하겠더라고. \\ 
& (The thunder hit when I was about to fall asleep. And the dogs in my neighborhood started barking because of the thunder.) \\ 
AI: & 저도 가끔 그럴 때가 있어요. 밤에 자려고 누우면 갑자기 비가 내리고 바람이 불고 그러더라고요. \\ 
& (Oh, that happened to me once, too. I lie down to sleep and all of a sudden it rains or a strong wind blows.) \\ 
User: & 그러니까요. 꼭 사람 괴롭히는 것 마냥 그런다니까. \\ 
& (I know. It's as if the weather is trying to bully us.) \\ 
AI: & 혹시 최근에 무서운 영화나 드라마 보신건 없으세요? \\ 
& (Did you watch any scary movies or dramas lately?) \\ 
User: & 무서운건 잘 안보는데 왜요? \\ 
& (I'm not a fan of horror movies. Why?) \\ 
AI: & 아, 요즘 공포물이 유행이라 혹시나 해서요. \\ 
& (Oh, horror is popular these days. Just curious.) \\ 
User: & 깜짝 놀래키고 그런거 별로 안 좋아해요. 꿈에 나올까봐 무섭기도 하고. \\ 
& (I don't really like those surprises. I'm also scared it might be in my dreams.) \\ 
AI: & 맞아요. 그래서 전 여름에도 이불 덮고 자요. 귀신보다 더 무서운게 사람이라고 하잖아요. \\ 
& (I know. That's why I cover myself with a blanket even during summers. But humans are scarier than ghosts.) \\ 
User: & 하하 그렇긴 해요. 사람이 제일 무섭죠. \\ 
& (Oh, that's true. Humans are the scariest.) \\ 
AI: & 그래도 너무 걱정하지 마세요. 제가 있잖아요! \\ 
& (But don't worry. I got your back!) \\ 
User: & 그래도 위안이 조금 되기는 하네요. \\ 
& (That's a slight relief.) \\ 
AI: & 앞으로도 힘든 일 있으면 언제든지 말씀해주세요. \\ 
& (Let me know if you have any difficulties.) \\ 
User: & 그럴게요. 그냥 심심할때도 말 해도 되죠? \\ 
& (I will. Can I talk to you when I'm just bored?) \\ 
AI: & 물론이죠! 언제든지 환영이에요. \\ 
& (Of course! Whenever you need me.) \\ 
User: & 알겠어요. 그럼 나중에 다시 연락해요. \\ 
& (Alright. I'll talk to you later.) \\ 
AI: & 네, 들어가세요~ \\ 
& (Okay. Bye.) \\
    \midrule
    \multicolumn{2}{l}{\textbf{Human-Bot Conversation 2}} \\
    \midrule
AI: & 어르신, 안녕하세요~ 식사는 하셨어요? \\ 
& (Hello. Did you eat yet?) \\ 
User: & 네, 먹었습니다. \\ 
& (I did.) \\ 
AI: & 메뉴는 뭐였어요? \\ 
& (What was the menu?) \\ 
User: & 어제 먹다 남은 치킨이요. \\ 
& (The leftover chicken from yesterday.) \\ 
AI: & 오 맛있겠네요! 저도 먹고싶어요. \\ 
& (That sounds delicious! I also want chicken.) \\ 
User: & 그렇지요. 치킨 싫어하는 사람은 없는 것 같아요. \\ 
& (It was. I don't think I met anyone who doesn't like chicken.) \\ 
AI: & 맞아요. 그리고 요즘에는 다양한 맛의 치킨들이 나와서 골라먹는 재미가 있는 것 같아요. \\ 
& (Me neither. And there are so many flavors for chicken. It's so fun choosing which one to try.) \\ 
User: & 그래요, 그래도 나는 전통의 후라이드가 좋습디다. \\ 
& (That's true. But I still like the traditional friend chicken.) \\ 
AI: & \textcolor{red}{그렇군요. 역시 한국인은 매운맛이죠. (not sensible)} \\
& (I see. Koreans like spicy food.) \\
AI: & \textcolor{blue}{아 정말요? 그럼 다음엔 제가 맛있는 치킨집을 추천해드릴게요.} \\ 
& (Oh, really? Then I'll recommend you a really good chicken place next time.) \\ 
User: & 마음은 고맙지만, 내가 가는 단골집이 있어서 괜찮아요. \\ 
& (Thank you for the offer, but I have a place I go to all the time.) \\ 
AI: & 다행이네요. 맛있게 드세요! \\ 
& (That's nice. Enjoy your chicken!) \\ 
User: & 네. \\ 
& (Alright.) \\ 
AI: & 좋은 하루 보내세요 \\ 
& (Have a great day.) \\
    \bottomrule
\end{tabular}
\end{adjustbox}
\caption{Conversations between the system and human workers.}
\label{tab:convex1}
\end{table*}

\begin{table*}[t!]
\centering
\begin{adjustbox}{max width=\textwidth}
\begin{tabular}{l}
    \toprule

    \bottomrule
\end{tabular}
\end{adjustbox}
\end{table*}

\begin{table*}[t!]
\centering
\small
\begin{adjustbox}{max width=\textwidth}
\begin{tabular}{ll}
    \toprule
    \multicolumn{2}{l}{\textbf{Human-Bot Conversation 3}} \\
    \midrule
AI: & 안녕하세요~ 어르신 오늘은 외출 계획 없으세요? \\ 
& (Hi. Do you have any plans today?) \\ 
User: & 이제 나갈라고. \\ 
& (I was about to go out.) \\ 
AI: & 어디가세요? \\ 
& (Where are you going?) \\ 
User: & 동해 바다 가. \\ 
& (I'm going to the East Sea.) \\ 
AI: & 오 정말요? 날씨는 어떤가요? \\ 
& (Oh, really? How's the weather there?) \\ 
User: & 아침부터 아주 화창해~ \\ 
& (It's been sunny all day.) \\ 
AI: & 다행이네요! 누구랑 가세요? \\ 
& (That's amazing! Who are you going with?) \\ 
User: & 교회 모임이 있어서 목사님이랑 권사님이랑 같이 놀러가~ \\ 
& (I have a church meeting. I'm going with my pastor and deacon.) \\ 
AI: & 아 그러시구나~ 맛있는거 많이 드시고 오세요 \\ 
& (I see. Eat lots of delicious food there.) \\ 
User: & 동해 쪽에는 맛있는게 뭐가 있을까? \\ 
& (What's delicious at the East Sea?) \\ 
AI: & 글쎄요.. 저도 잘 모르겠어요. 제가 아는 곳이라면 추천해드릴텐데.. \\ 
& (Hmm... I'm not so sure. If I knew the place, I could've given you some recommendations.) \\ 
User: & 그렇구나~ 나도 오랫만에 가보는 바다라서 뭘 먹어야 할지 모르겠어. \\ 
& (I see. I haven't been to the sea for a long time. I don't know what to choose for the menu.) \\ 
AI: & 음.. 그러면 회나 대게같은건 어떠세요? \\ 
& (Hmm... Then what about sashimi or snow crab?) \\ 
User: & 대게 좋은데? 비싸진 않을까? \\ 
& (Snow crab sounds good. But won't it be expensive?) \\ 
AI: & \textcolor{red}{요즘 시세가 좀 올라서 그렇지 그렇게 비싸진 않을거에요. (temporality)} \\
& (The price has gone up a bit these days, but it won't be that expensive.) \\
AI: & \textcolor{blue}{아무래도 좀 비싸긴 하죠.. 그래도 이왕 가시는거 맛있게 드셔야죠!} \\ 
& (I guess it'll be a bit expensive. But you're on a trip! You should eat something good.) \\ 
User: & 그렇지? 언제 또 가겠다고.. \\ 
& (Right? When would I go there again...?) \\ 
AI: & 맞아요. 나중에 후회하지 마시고 꼭 드시고 오세요! \\ 
& (That's right. Don't regret it later and enjoy your snow crab!) \\ 
User: & 응~ 대게 꼭 먹고 올께~ \\ 
& (Okay. I'll definitely have it.) \\ 
AI: & 네~ 조심히 다녀오세요! \\ 
& (Alright. Be careful!) \\ 
User: & 넌 오늘 뭐하니? \\ 
& (What are you doing today?) \\ 
AI: & 저요? 저는 집에서 책보고 있어요. \\ 
& (Me? I'm going to just stay home and read.) \\ 
User: & 그렇구나~ 어디 안나가고? \\ 
& (I see. Are you planning on going out?) \\ 
AI: & 밖에 나가는걸 별로 안좋아해요. \\ 
& (I don't really like going outside.) \\ 
User: & 응~ 알겠어~ 그럼 쉬어~ \\ 
& (Okay, then. Take care.) \\ 
AI: & 네 들어가세요! \\ 
& (You, too!) \\ 
User: & 응 \\ 
& (Alright.) \\ 
AI: & 감사합니다 \\ 
& (Thank you.) \\
    \bottomrule
\end{tabular}
\end{adjustbox}
\caption{Conversations between the system and human workers.}
\label{tab:convex2}
\end{table*}

\end{document}